\documentclass[10pt,twocolumn,letterpaper]{article}

\usepackage{cvpr}
\usepackage{times}
\usepackage{epsfig}
\usepackage{graphicx}
\usepackage{amsmath}
\usepackage{amssymb}
\usepackage{subfigure}
\usepackage{floatrow}
\usepackage{cite}
\usepackage{tabularx}
\usepackage[table,xcdraw]{xcolor}
\usepackage{multirow}
\usepackage{color}
\usepackage{algorithm}
\usepackage{amsthm}
\usepackage{booktabs}
\usepackage{subfig}
\usepackage{lineno}
\usepackage{colortbl}
\usepackage[]{algpseudocode}
\usepackage{floatrow}
\usepackage{authblk}
\newcommand{\RN}[1]{\textup{\uppercase\expandafter{\romannumeral#1}}}
\usepackage[pagebackref=true,breaklinks=true,letterpaper=true,colorlinks,bookmarks=false]{hyperref}

 \cvprfinalcopy 

\def\cvprPaperID{2866} 

\ifcvprfinal\fi
\begin{document}

\title{Deeply Aggregated Alternating Minimization for Image Restoration}

\author[1]{Youngjung Kim}
\author[1]{Hyungjoo Jung}
\author[2]{Dongbo Min}
\author[1]{Kwanghoon Sohn}
\affil[1]{Yonsei University}\affil[2]{Chungnam National University}
\maketitle

\begin{abstract}
Regularization-based image restoration has remained an active research topic in computer vision and image processing.
It often leverages a guidance signal captured in different fields as an additional cue.
In this work, we present a general framework for image restoration, called deeply aggregated alternating minimization (DeepAM).
We propose to train deep neural network to advance two of the steps in the conventional AM algorithm: proximal mapping and $\beta$-continuation.
Both steps are learned from a large dataset in an end-to-end manner.
The proposed framework enables the convolutional neural networks (CNNs) to operate as a prior or regularizer in the AM algorithm.
We show that our learned regularizer via deep aggregation outperforms the recent data-driven approaches as well as the nonlocal-based methods.
The flexibility and effectiveness of our framework are demonstrated in several image restoration tasks, including single image denoising, RGB-NIR restoration, and depth super-resolution.
\end{abstract}

\section{Introduction}
Image restoration is a process of reconstructing a clean image from a degraded observation.
The observed data is assumed to be related to the ideal image
through a forward imaging model that accounts for noise, blurring,
and sampling. However, a simple modeling only with the observed data
is insufficient for an effective restoration, and thus a
priori constraint about the solution is commonly used. To this end,
the image restoration is usually formulated as an energy
minimization problem with an explicit regularization function (or regularizer).
Recent work on joint restoration leverages a guidance signal,
captured from different devices, as an additional cue to regularize
the restoration process.
These approaches have been successfully applied to various
applications including joint upsampling \cite{Ferstl2013},
cross-field noise reduction \cite{Shen20152}, dehazing
\cite{Shen2015}, and intrinsic image decomposition \cite{Chen2013}.

The regularization-based image restoration involves the
minimization of non-convex and non-smooth energy functionals for
yielding high-quality restored results. Solving such functionals
typically requires a huge amount of iterations, and thus an
efficient optimization is preferable, especially in some
applications the runtime is crucial. One of the most popular
optimization methods is the alternating minimization (AM) algorithm
\cite{Wang2008} that introduces auxiliary variables. The energy
functional is decomposed into a series of subproblems that is
relatively simple to optimize, and the minimum with respect to each
of the variables is then computed.
For the image restoration, the AM algorithm has been widely adopted
with various regularization functions, e.g., total variation
\cite{Wang2008}, $L_0$ norm \cite{Xu2011}, and $L_p$ norm
(hyper-Laplacian) \cite{Krishnan09}. It is worth noting that
these functions are all handcrafted models.
The hyper-Laplacian of image gradients \cite{Krishnan09} reflects
the statistical property of natural images relatively well, but the
restoration quality of gradient-based regularization methods using
the handcrafted model is far from that of the state-of-the-art
approaches \cite{Schmidt2014,Chen2015}. In general, it is
non-trivial to design an optimal regularization function for a
specific image restoration problem.

Over the past few years, several attempts have been made to overcome
the limitation of handcrafted regularizer by
learning the image restoration model from a large-scale training data
\cite{Schmidt2014,Chen2015,Zoran2011}. In this work, we propose a
novel method for image restoration that effectively uses a
data-driven approach in the energy minimization framework, called
\emph{deeply aggregated alternating minimization} (DeepAM). Contrary
to existing data-driven approaches that just produce the restoration
results from the convolutional neural networks (CNNs), we design the
CNNs to implicitly learn the regularizer of the AM algorithm.
Since the CNNs are fully integrated into the
AM procedure, the whole networks can be learned simultaneously in
an end-to-end manner. We show that our simple model learned from
the deep aggregation achieves better results than the recent
data-driven approaches \cite{Schmidt2014,Chen2015,Li2016} as well as the state-of-the-art
nonlocal-based methods \cite{Dabov2007,Gu2014}.

Our main contributions can be summarized as follows:
\begin{itemize}
\renewcommand\labelitemi{\tiny$\bullet$}
  \item We design the CNNs to learn the regularizer of the AM algorithm, and train the whole networks in an end-to-end manner.
  \vspace{-5pt}
  \item We introduce the aggregated (or multivariate) mapping in the AM algorithm, which leads to a better restoration model than the conventional point-wise proximal mapping.
  \vspace{-5pt}
  \item We extend the proposed method to joint restoration tasks. It has broad applicability to a variety of restoration problems, including image denoising, RGB/NIR restoration, and depth super-resolution.
\end{itemize}

\section{Related Work}
\paragraph{Regularization-based image restoration}
Here, we provide a brief review of the regularization-based image
restoration. The total variation (TV) \cite{Wang2008} has been
widely used in several restoration problems thanks to its convexity
and edge-preserving capability. Other regularization functions such
as total generalized variation (TGV) \cite{Bredies2010} and $L_p$
norm \cite{Krishnan09} have also been employed to penalize an image
that does not exhibit desired properties. Beyond these handcrafted
models, several approaches have been attempted to learn the regularization
model from training data \cite{Schmidt2014,Chen2015}. Schmidt
\emph{et al.} \cite{Schmidt2014} proposed a cascade of shrinkage
fields (CSF) using learned Gaussian RBF kernels. In \cite{Chen2015},
a nonlinear diffusion-reaction process was modeled by using
parameterized linear filters and regularization functions. Joint
restoration methods using a guidance image captured under different
configurations have also been studied \cite{Agrawal05,Ferstl2013,Shen2015,Li2016}. In \cite{Agrawal05},
an RGB image captured in dim light was restored using flash and
non-flash pairs of the same scene. In \cite{Ferstl2013,Kim20162},
RGB images was used to assist the regularization process of a
low-resolution depth map. Shen \emph{et al.} \cite{Shen2015}
proposed to use dark-flashed NIR images for the restoration of noisy
RGB image. Li \emph{et al.} used the CNNs to selectively transfer
salient structures that are consistent in both guidance and target
images \cite{Li2016}.

\paragraph{Use of energy minimization models in deep network}
The CNNs lack imposing the regularity constraint on adjacent similar
pixels, often resulting in poor boundary localization and spurious
regions. To deal with these issues, the integration of energy
minimization models into CNNs has received great attention
\cite{Ranftl2014,Zheng2015,Riegler2016,Riegler20162}. Ranftl
\emph{et al.} \cite{Ranftl2014} defined the unary and pairwise terms
of Markov Random Fields (MRFs) using the outputs of CNNs, and
trained network parameters using the bilevel optimization.
Similarly, the mean field approximation for fully connected
conditional random fields (CRFs) was modeled as recurrent neural networks (RNNs) \cite{Zheng2015}.
A nonlocal Huber regularization was combined
with CNNs for a high quality depth restoration \cite{Riegler2016}.
Riegler \emph{et al.} \cite{Riegler20162} integrated anisotropic TGV
into the top of deep networks. They also formulated
the bilevel optimization problem and trained the network in an
end-to-end manner by unrolling the TGV minimization. Note that the
bilevel optimization problem is solvable only when the energy
minimization model is convex and is twice differentiable
\cite{Ranftl2014}. The aforementioned methods try to integrate
handcrafted regularization models into top of the CNNs. In
contrast, we design the CNNs to parameterize the regularization
process in the AM algorithm.

\begin{figure}[!]
\centering
\subfigure{\includegraphics[width=0.179\textheight]{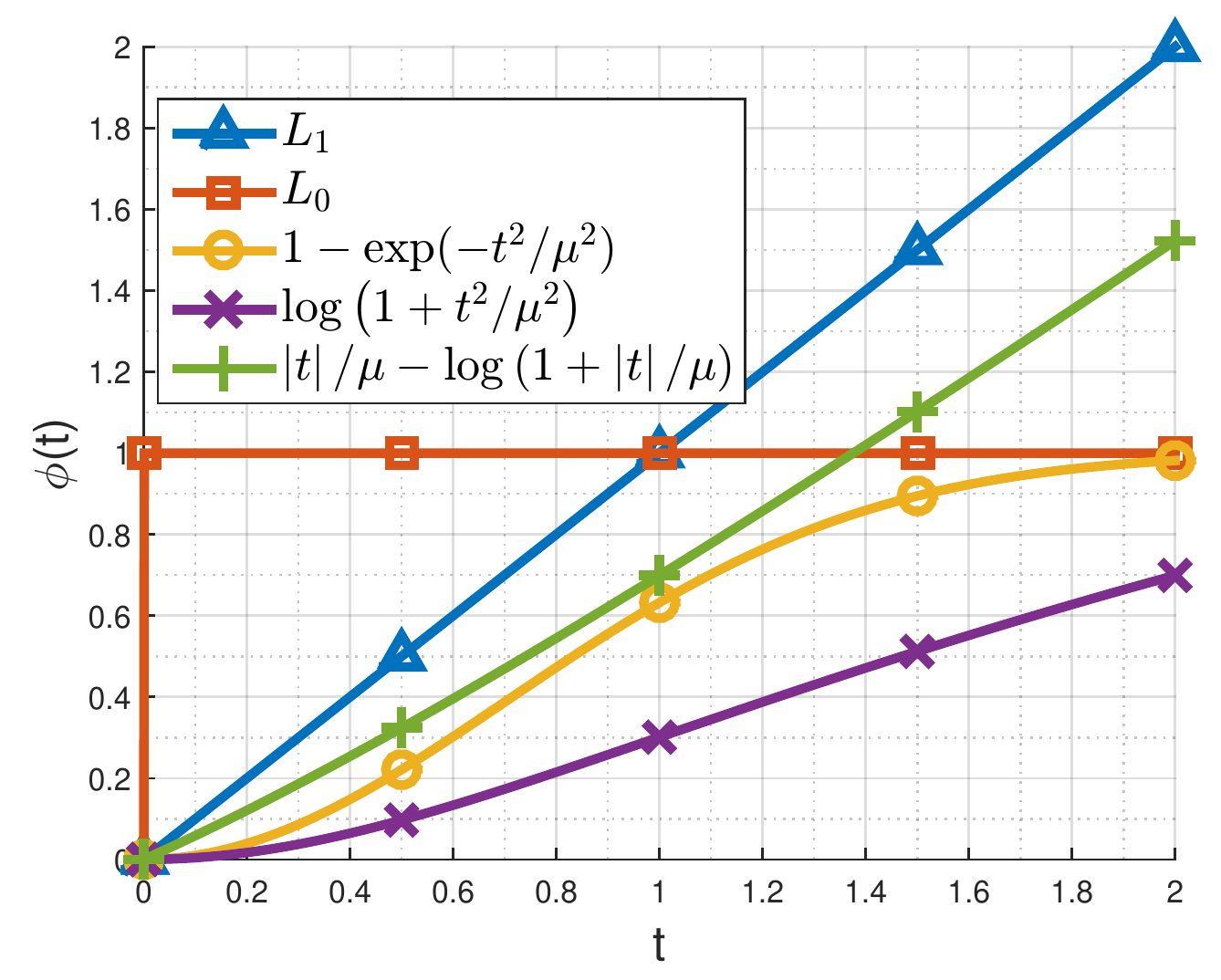}}
\subfigure{\includegraphics[width=0.186\textheight]{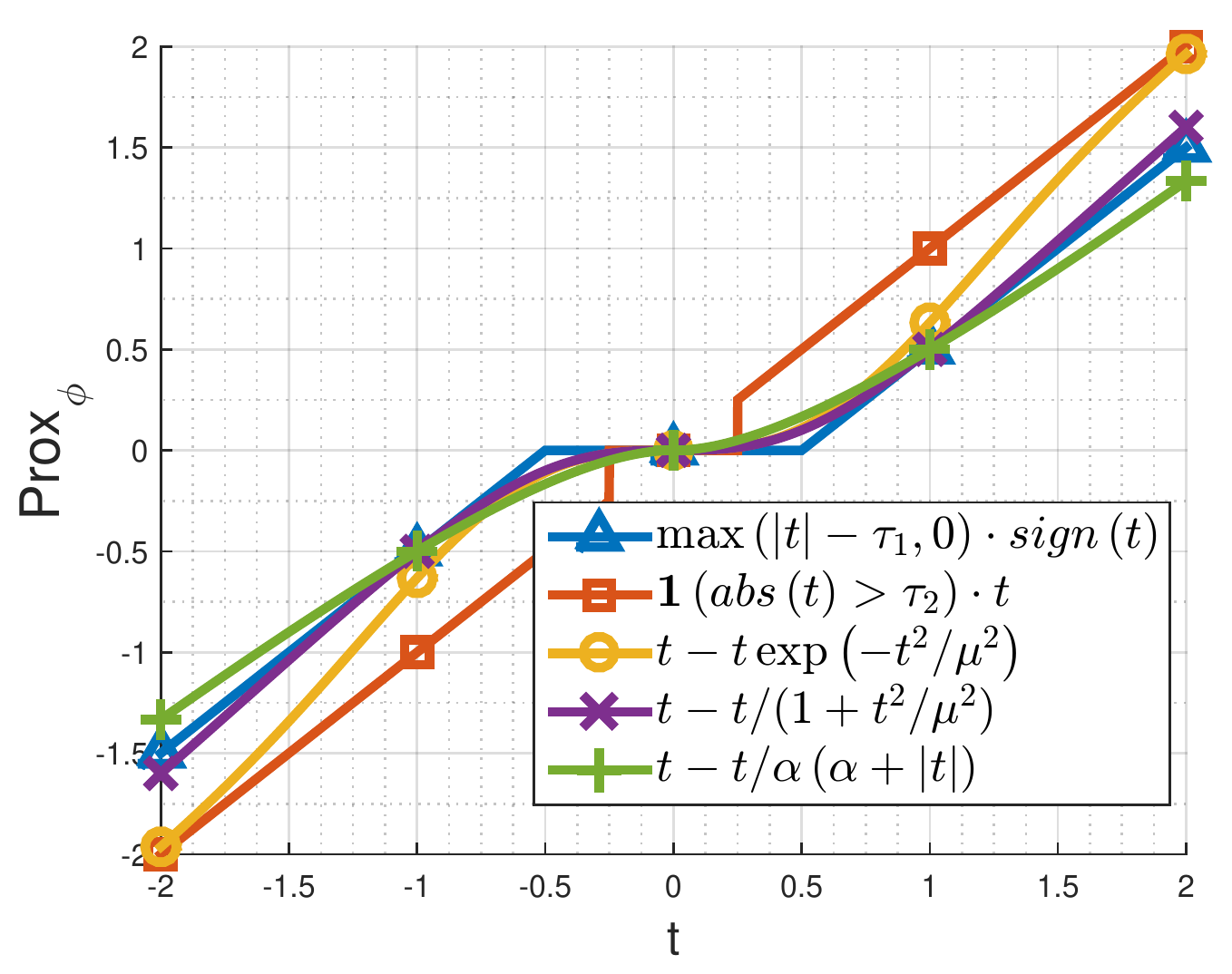}}
\vspace{-15pt}
\caption{Illustrations of the regularization function $\Phi$ (left) and the corresponding proximal mapping (right). The main purpose of this mapping is to remove $D{u^k}$ with a small magnitude, since they are assumed to be caused by noise. Instead of such handcrafted regularizers, we implicitly parameterize the regularization function using the deep aggregation, leading to a better restoration algorithm.
\vspace{-5pt}}
\label{fig:aa}
\end{figure}

\section{Background and Motivation}
The regularization-based image reconstruction is a powerful
framework for solving a variety of inverse problems in computational
imaging. The method typically involves formulating a data term for
the degraded observation and a regularization term for the image to
be reconstructed. An output image is then computed by minimizing an
objective function that balances these two terms. Given an observed
image $f$ and a balancing parameter $\lambda$, we solve the
corresponding optimization problem\footnote{For the
super-resolution, we treat $f$ as the bilinearly upsampled image
from the low-resolution input.}:

\begin{equation}
\label{eq:1}
\mathop {\arg \min }\limits_u \frac{\lambda }{2}{\left\| {u - f} \right\|^2} + \Phi (Du).
\end{equation}
$D{u}$ denotes the $[{D_x}{u},{D_y}{u}]$, where $D_x$ (or $D_y$) is
a discrete implementation of $x$-derivative (or $y$-derivative) of
the image. $\Phi$ is a regularization function
that enforces the output image $u$ to meet desired statistical
properties. The unconstrained optimization problem of (\ref{eq:1}) can be
solved using numerous standard algorithms. 
In this paper, we focus on the additive form of alternating
minimization (AM) method \cite{Wang2008}, which is the ad-hoc for a
variety of problems in the form of (\ref{eq:1}).

\begin{figure*}[!]
\centering
\subfigure[Noisy input]{\includegraphics[width=0.15\textheight]{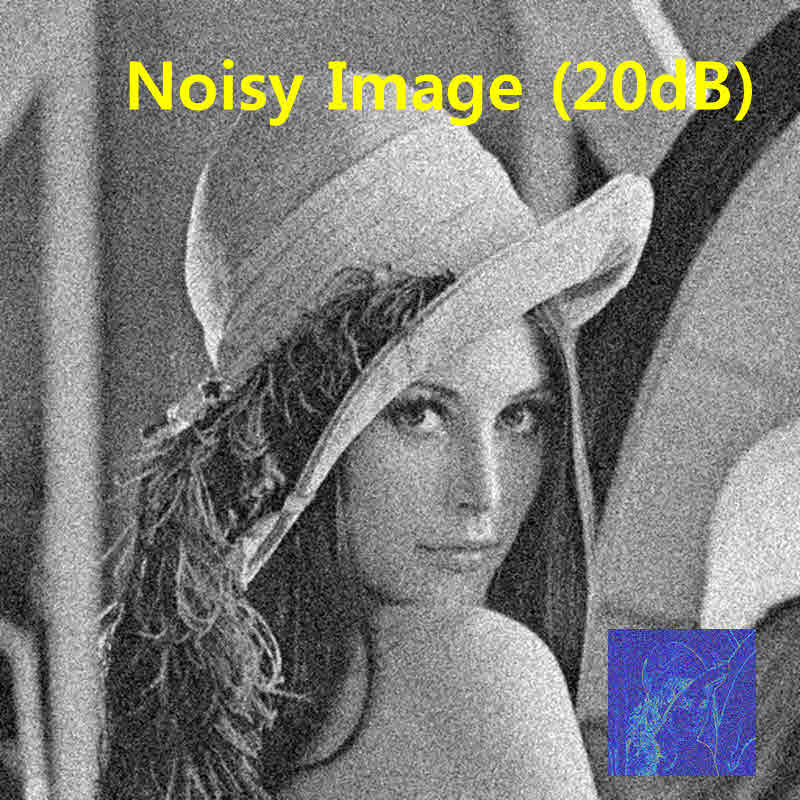}}
\subfigure[TV \cite{Wang2008}]{\includegraphics[width=0.15\textheight]{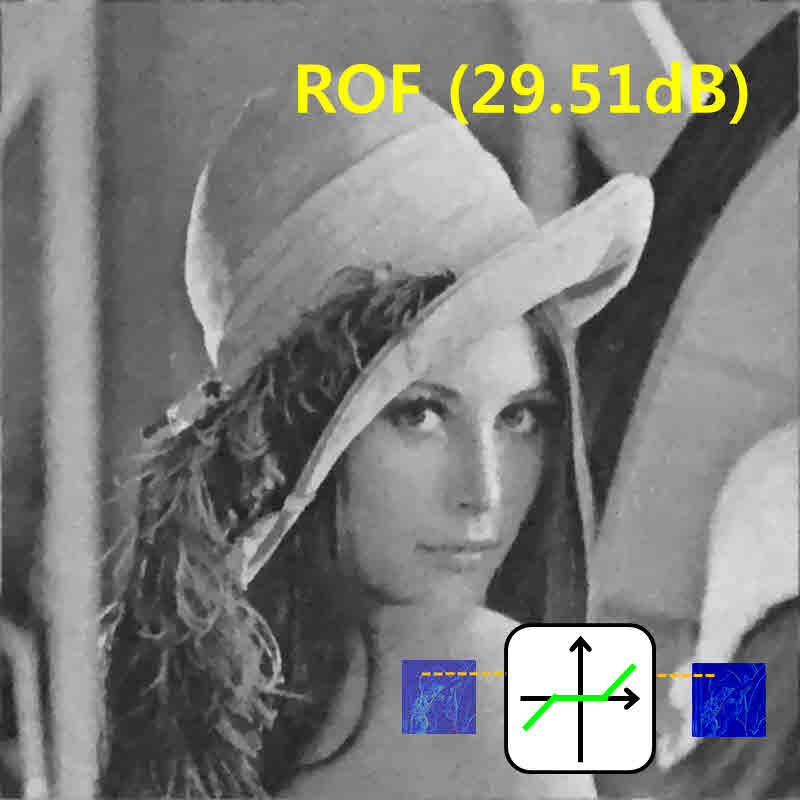}}
\subfigure[CSF \cite{Schmidt2014}]{\includegraphics[width=0.15\textheight]{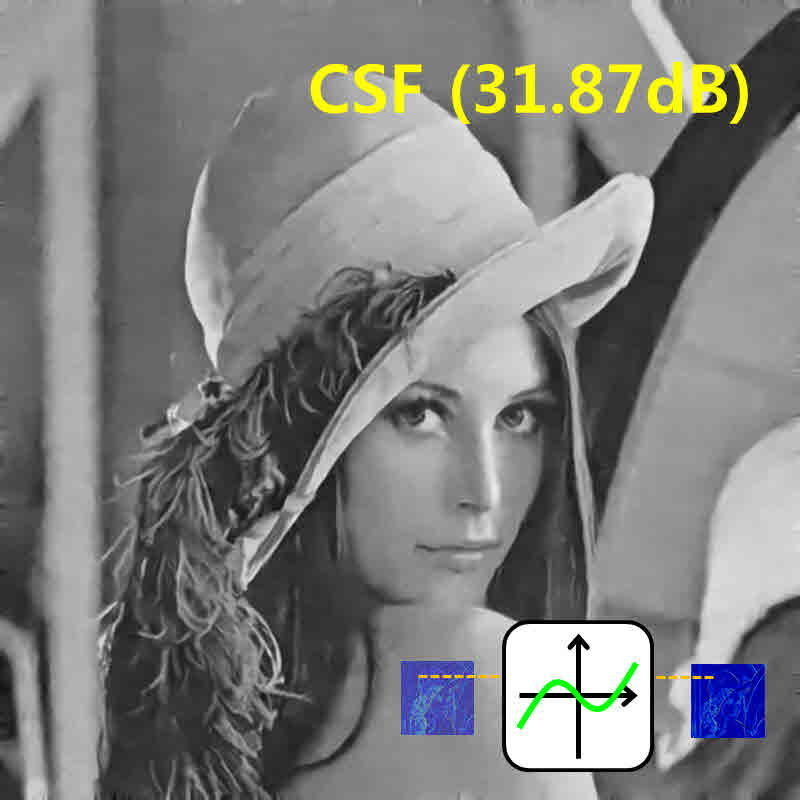}}
\subfigure[Ours]{\includegraphics[width=0.15\textheight]{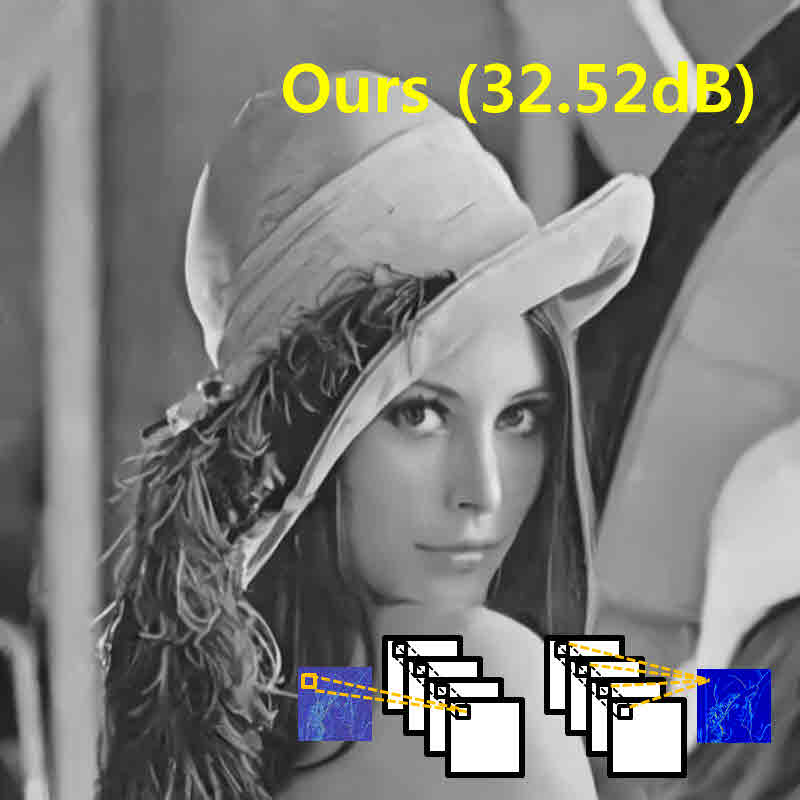}}
\subfigure[Reference]{\includegraphics[width=0.15\textheight]{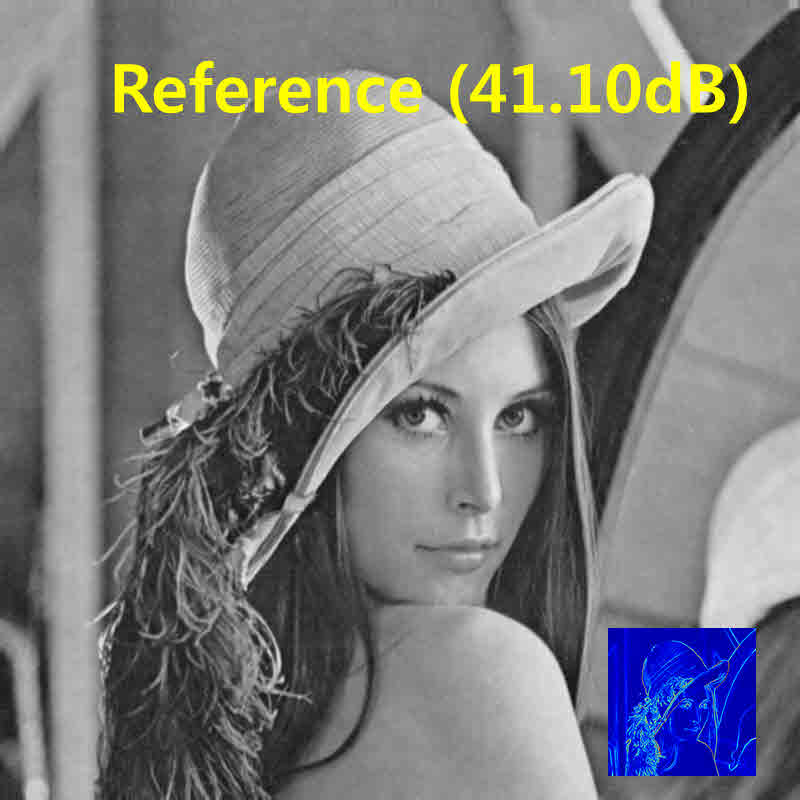}}\\
\vspace{-5pt}
\caption{Examples of single image denoising: (a) input image, (b) TV \cite{Wang2008}, (c) CSF \cite{Schmidt2014}, and (d) ours. (e) is obtained after one step of the AM iteration using $D{u^ * }$ with $\lambda=5$, where $u^*$ is a noise-free image. Our deeply aggregated AM outperforms existing point-wise mapping operators.}
\label{fig:bb}
\end{figure*}

\subsection{Alternating Minimization}
The idea of AM method is to decouple the data and regularization
terms by introducing a new variable $v$ and to reformulate (\ref{eq:1}) as
the following constrained optimization problem:

\begin{equation}
\label{eq:2}
\mathop {\min }\limits_{u,v} \frac{\lambda }{2}{\left\| {u - f} \right\|^2} + \Phi (v) ,\; {\rm{subject}}\;{\rm{to}}\;\;v = D{u}.
\end{equation}
We solve (\ref{eq:2}) by using the penalty technique \cite{Wang2008},
yielding the augmented objective function.

\begin{equation}
\label{eq:3}
\mathop {\min }\limits_{u,v} \frac{\lambda }{2}{\left\| {u - f} \right\|^2} + \Phi (v) + \frac{\beta }{2}{\left\| {Du - v} \right\|^2},
\end{equation}
where $\beta$ is the penalty parameter. The AM algorithm consists of repeatedly performing the following steps until convergence.

\begin{equation}
\label{eq:4}
\begin{array}{*{20}{l}}
{{v^{k + 1}} = \mathop {\arg \min }\limits_v \Phi (v) + \frac{{{\beta ^k}}}{2}{{\left\| {D{u^{k}} - v} \right\|}^2},}\\
{{u^{k + 1}} = \mathop {\arg \min \frac{\lambda }{2}}\limits_u {{\left\| {u - f} \right\|}^2} + \frac{{{\beta ^k}}}{2}{{\left\| {Du - {v^{k+1}}} \right\|}^2},}\\
{{\beta ^{k + 1}} = \alpha {\beta ^k},}
\end{array}
\end{equation}
where $\alpha>1$ is a continuation parameter. When $\beta$ is large enough, the variable $v$ approaches $D{u}$, and thus (\ref{eq:3}) converges to the original formulation (\ref{eq:1}).

\subsection{Motivation}
Minimizing the first step in (\ref{eq:4}) varies depending on the choices of
the regularization function $\Phi$ and $\beta$. This step can be regarded as the
proximal mapping \cite{Parikh2014} of $D{u^{k}}$ associated with
$\Phi$. When $\Phi$ is the sum of $L_1$ or $L_0$ norm, it amounts to
soft or hard thresholding operators (see Fig.~\ref{fig:aa} and
\cite{Parikh2014} for various examples of this relation). Such
mapping operators may not unveil the full potential of the
optimization method of (\ref{eq:4}), since $\Phi$ and $\beta$ are chosen
manually. Furthermore, the mapping operator is performed for each
pixel individually, disregarding spatial correlation with
neighboring pixels.

Building upon this observation, we propose the new approach  in
which the regularization function $\Phi$ and the penalty parameter
$\beta$ are learned from a large-scale training dataset. Different
from the point-wise proximal mapping based on the handcrafted
regularizer, the proposed method learns and aggregates the mapping
of $Du^k$ through CNNs.

\section{Proposed Method}
In this section, we first introduce the DeepAM for a single image
restoration, and then extend it to joint restoration tasks. In the
following, the subscripts $i$ and $j$ denote the location of a pixel
(in a vector form).

\begin{figure*}[!]
\centering
\subfigure{\includegraphics[width=0.8\textheight]{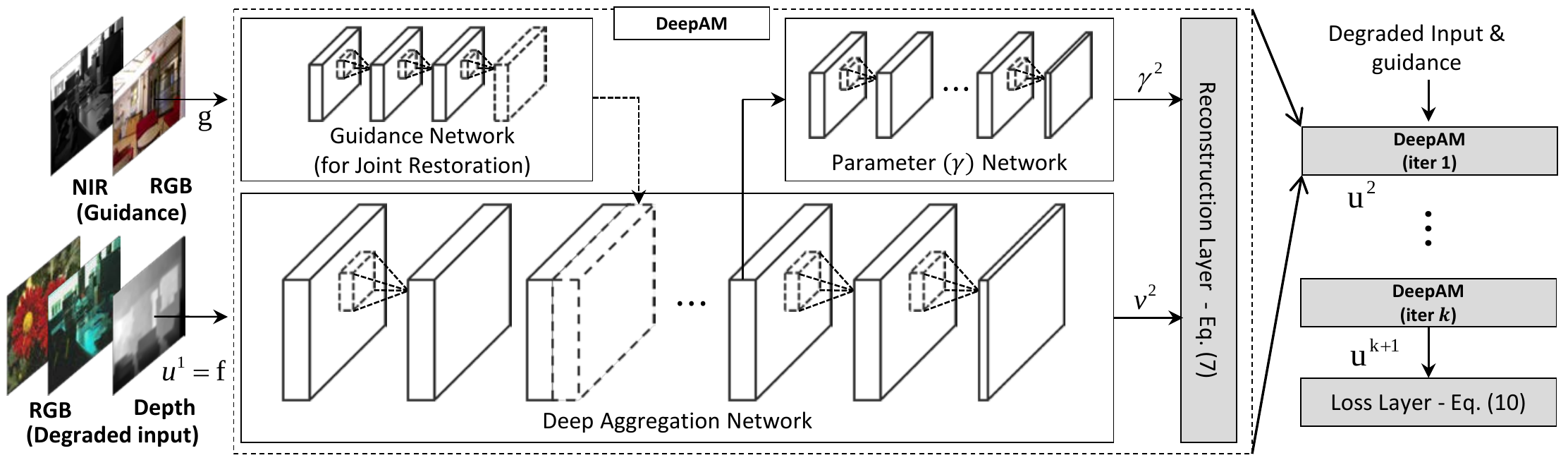}}\\
\vspace{-5pt}
\caption{One iteration of our model consists of four major components: deep aggregation network, guidance network, $\gamma$-parameter network, and reconstruction layer. The spatially varying $\gamma$ is estimated by exploiting features from intermediate layers of the deep aggregation network.
All of these sub-networks are cascaded by iterating (\ref{eq:5}) and (\ref{eq:6}), and the final output is then entered into the loss layer.
\vspace{-5pt}}
\label{fig:cc}
\end{figure*}

\subsection{Deeply Aggregated AM}
We begin with some intuition about why our learned and aggregated
mapping is crucial to the AM algorithm. The first step in (\ref{eq:4}) maps
$D{u^{k}}$ with a small magnitude into zero since it is assumed
that they are caused by noise, not an original signal.
Traditionally, this mapping step has been applied in a point-wise
manner, not to mention whether it is learned or not. With $\Phi (v)
= \sum\nolimits_i {\phi ({v_i})}$, Schmidt \emph{et al.}
\cite{Schmidt2014} modeled the point-wise mapping function as
Gaussian RBF kernels, and learned their mixture
coefficients\footnote{When $\Phi (v) = \sum\nolimits_i {\phi
({v_i})}$, the first step in (\ref{eq:4}) is separable with respect to each
$v_i$. Thus, it can be modeled by point-wise operation.}.
Contrarily, we do not presume any property of $\Phi$. We instead
train the multivariate mapping process ($D{u^{k}} \to {v^{k + 1}}$)
associated with $\Phi$ and $\beta$ by making use of the CNNs.
Figure~\ref{fig:bb} shows the denoising examples of TV \cite{Wang2008}, CSF
\cite{Schmidt2014}, and ours. Our method outperforms
other methods using the point-wise mapping based on handcrafted model (Fig.~\ref{fig:bb}(b)) or learned model (Fig.~\ref{fig:bb}(c)) (see the insets).

We reformulate the original AM iterations in (\ref{eq:4}) with the following
steps\footnote{The gradient operator $D$ is absorbed into the CNNs.}.

\begin{equation}
\label{eq:5}
\left( {{v^{k + 1}},{\gamma ^{k + 1}}} \right)  \Leftarrow  {{\cal D}_{CNN}}({u^k},{w^k_{u}}),
\end{equation}
\vspace{-5pt}
\begin{equation}
\label{eq:6}
{u^{k + 1}} = \mathop {\arg \min }\limits_u {\small\| {{\Gamma ^{k + 1}}(u - f)} \small\|^2} + {\small\| {Du - {v^{k + 1}}} \small\|^2},
\end{equation}
where ${D_{CNN}}( \cdot ,w_u^k)$ denotes a convolutional network
parameterized by $w_u^k$ and $\Gamma ^{k + 1}=diag(\gamma ^{k +
1})$.
Note that $\beta$ is completely absorbed into the CNNs, and fused with the balancing parameter $\gamma$ (which will also be learned).
$v^{k + 1}$ is estimated by deeply aggregating $u^k$ through CNNs.
This formulation allows us to turn the
optimization procedure in (\ref{eq:1}) into a cascaded neural network
architecture, which can be learned by the standard back-propagation
algorithm \cite{Mozer1989}.

The solution of (\ref{eq:6})
satisfies the following linear system:

\begin{figure}[]
\subfigure[$u^1$]{\includegraphics[width=0.12\textheight]{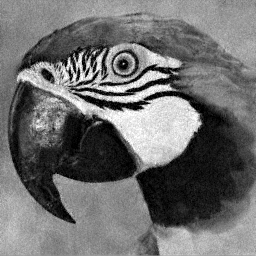}}
\subfigure[$u^2$]{\includegraphics[width=0.12\textheight]{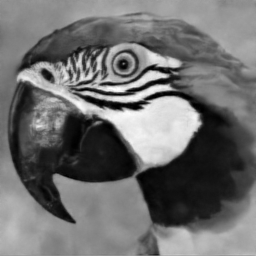}}
\subfigure[$u^3$]{\includegraphics[width=0.12\textheight]{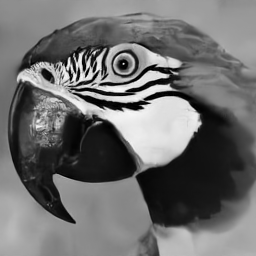}}\\
\vspace{-3pt}
\subfigure[$\lVert v^1 \rVert$]{\includegraphics[width=0.12\textheight]{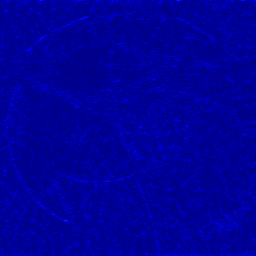}}
\subfigure[$\lVert v^2 \rVert$]{\includegraphics[width=0.12\textheight]{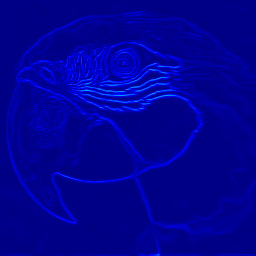}}
\subfigure[$\lVert v^3 \rVert$]{\includegraphics[width=0.12\textheight]{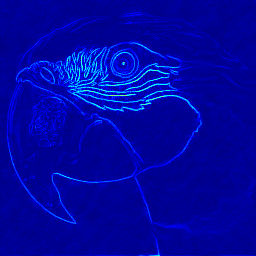}}\\
\vspace{-5pt}
\caption{The denoising results obtained by our DeepAM (trained with $K=3$ iterations in Fig~\ref{fig:cc}).
See the text for details.
\vspace{-10pt}}
\label{fig:dd}
\end{figure}

\begin{equation}
\label{eq:7}
L{u^{k + 1}} = {\Gamma ^{k + 1}}f + {D^T}{v^{k + 1}},
\end{equation}
where the Laplacian matrix $L = ({\Gamma ^{k + 1}} + {D^T}D)$. It
can be seen that (\ref{eq:7}) plays a role of naturally imposing the spatial
and appearance consistency on the intermediate output image
$u^{k+1}$ using a kernel matrix $A_{ij}=L^{-1}_{ij}$
\cite{Zheng2015}. The linear system of (\ref{eq:7}) becomes the part of deep
neural network (see Fig.~\ref{fig:cc}). When $\gamma$ is a constant, the
block Toeplitz matrix $L$ is diagonalizable with the fast
Fourier transform (FFT).
However, in our framework, the direct application of FFT is
not feasible since $\gamma$ is spatially varying for the adaptive
regularization. Fortunately, the matrix $L$ is still
sparse and positive semi-definite as the simple gradient operator
$D$ is used. We adopt the preconditioned conjugate gradient (PCG)
method to solve the linear system of (\ref{eq:7}). The incomplete Cholesky
factorization \cite{Suite} is used for computing the preconditioner.

Very recently, Chan \emph{et al.} \cite{Chan2016} replaced the
proximal mapping in (\ref{eq:4}) with an off-the-shelf image denoising
algorithm $\mathcal{D}_\sigma$, e.g., nonlocal means \cite{Buades2005}, as follows:

\begin{equation}
\label{eq:8}
{v^{k + 1}}  \Leftarrow  {\mathcal{D}_\sigma }(D{u^{k+1}}).
\end{equation}
Although this is conceptually similar to our aggregation
approach\footnote{Aggregation using neighboring pixels are commonly
used in state-of-the-arts denoising methods.}, the operator
$\mathcal{D}_\sigma$ in \cite{Chan2016} still relies on the
handcrafted model.
Figure~\ref{fig:cc} shows the proposed learning model for image restoration
tasks. The DeepAM, consisting of deep aggregation network, $\gamma$-parameter network, guidance
network (which will be detailed in next section), and reconstruction
layer, is iterated $K$ times, followed by the loss layer.

Figure~\ref{fig:dd} shows the denoising result of our method. Here, it is
trained with three passes of DeepAM. The input image is corrupted by
Gaussian noise with standard deviation $\sigma=25$. We can see that
as iteration proceeds, the high-quality restoration results are
produced. The trained networks in the first and second iterations
remove the noise, but intermediate results are over smoothed (Figs.~\ref{fig:dd}(a)
and (b)). The high-frequency information is then recovered in
the last network (Fig.~\ref{fig:dd}(c)). To analyze this behavior, let us date
back to the existing soft-thresholding operator, $v_i^{k + 1} = \max
\{ {\small|{Du^{k}}\small|_i} - 1/{\beta ^k}, 0\} {\text{sign}}{(Du)_i}$
in \cite{Wang2008}. The conventional AM method sets $\beta$ as a
small constant and increases it during iterations. When $\beta$ is
small, the range of $v$ is shrunk, penalizing large gradient
magnitudes. The high-frequency details of an image are recovered as $\beta$ increases.
Interestingly, the DeepAM shows very similar behavior (Figs.~\ref{fig:dd}(d)-(f)),
but outperforms the existing methods thanks to the aggregated mapping through the CNNs, as will be validated in experiments.

\subsection{Extension to Joint Restoration}
In this section, we extend the proposed method to joint restoration
tasks. The basic idea of joint restoration is to provide structural
guidance, assuming structural correlation between different kinds of
feature maps, e.g., depth/RGB and NIR/RGB. Such a constraint has
been imposed on the conventional mapping operator by considering
structures of both input and guidance images \cite{Kim20162}. Similarly, one can
modify the deeply aggregated mapping of (\ref{eq:5}) as follows:

\begin{equation}
\label{eq:9}
\left( {{v^{k + 1}},{\gamma ^{k + 1}}} \right)  \Leftarrow  {{\cal D}_{CNN}}({({u^k} \otimes  g)},{w^k_{u}}),
\end{equation}
where $g$ is a guidance image and $\otimes$ denotes a concatenation
operator. However, we find such early concatenation to be less
effective since the guidance image mixes heterogeneous data. This
coincides with the observation in the literature of multispectral
pedestrian detection \cite{Liu2016}. Instead, we adopt the halfway concatenation
similar to \cite{Liu2016,Li2016}. Another sub-network ${D_{CNN}}(g,w_g^k)$ is
introduced to extract the effective representation of the guidance
image, and it is then combined with intermediate features of
${D_{CNN}}(u^k,w_{u}^k)$ (see Fig.~\ref{fig:cc}).

\subsection{Learning Deeply Aggregated AM}
In this section, we will explain the network architecture and
training method using standard back-propagation algorithm. Our code
will be publicly available later.
\vspace{-5pt}
\paragraph{Network architecture}
One iteration of the proposed DeepAM consists of four major parts:
deep aggregation network, $\gamma$-parameter network, guidance network (for joint restoration), and reconstruction
layer, as shown in Fig.~\ref{fig:cc}. The deep aggregation network consists of
10 convolutional layers with $3\times3$ filters (a receptive field
is of $21\times21$). Each hidden layer of the network has 64 feature
maps. Since $v$ contains both positive and negative values, the
rectified linear unit (ReLU) is not used for the last layer.
The input distributions of all convolutional layers are normalized to
the standard Gaussian distribution \cite{Noh2015}.
The output channel of the deep aggregation network is 2 for
the horizontal and vertical gradients. We also extract the spatially
varying $\gamma$ by exploiting features from the eighth convolutional layer of
the deep aggregation network. The ReLU is used for ensuring the
positive values of $\gamma$.

For joint image restoration, the guidance network consists of 3
convolutional layers, where the filters operate on $3\times3$
spatial region. It takes the guidance image $g$ as an input, and
extracts a feature map which is then concatenated with the third convolutional
layer of the deep aggregation network. There are no parameters to be
learned in the reconstruction layer.

\begin{figure}
\begin{minipage}[]{0.58\textwidth}
\includegraphics[width=1\textwidth]{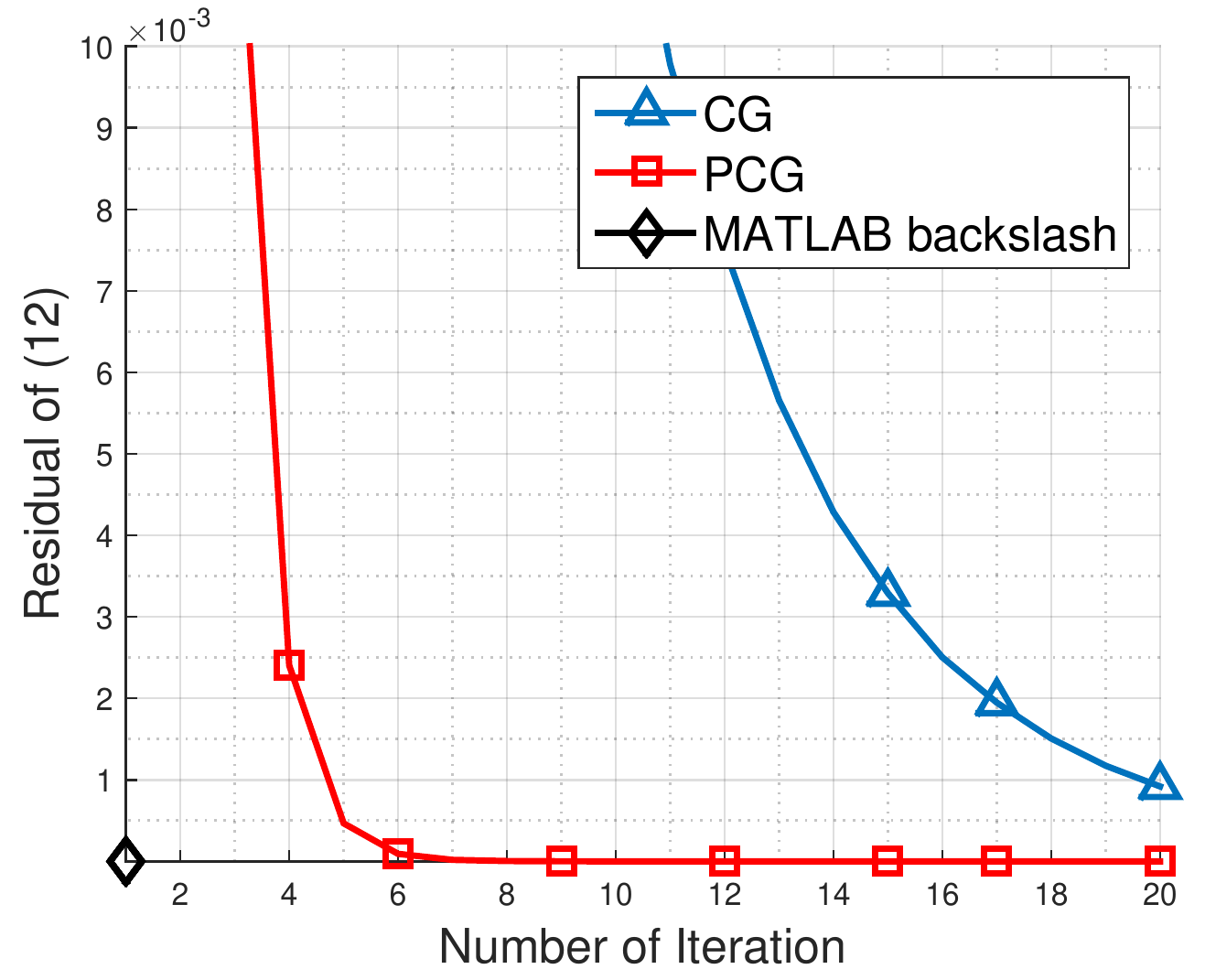}
\end{minipage}
\begin{minipage}[]{0.36\textwidth}
     \centering
     \begin{tabular}{cc}
     \toprule
     256$\times$256           & Time(s) \\ \midrule \midrule
     \begin{tabular}[c]{@{}c@{}}PCG\\ (10 iter) \end{tabular}      & 0.028   \\
     \begin{tabular}[c]{@{}c@{}}MATLAB\\ backslash \end{tabular}   & 0.155   \\
    \bottomrule
  \end{tabular}
\end{minipage}
\vspace{-3pt}
\caption{Figure in left shows the convergence of the PCG solver. A small number of PCG iterations are enough for the back-propagation.
The results of the MATLAB backslash is plotted in the origin.
The table in right compares the runtime of PCG with 10 iterations and direct MATLAB solver.
\vspace{-5pt}
}
\label{fig:ee}
\end{figure}
\vspace{-10pt}
\paragraph{Training}
The DeepAM is learned via standard back-propagation
algorithm \cite{Mozer1989}. We do not require the
complicated bilevel formulation \cite{Ranftl2014,Riegler20162}.
Given $M$ training image pairs $\{ {f^{(p)}},{g^{(p)}},{t^{(p)}}\} _{p =
1}^M$, we learn the network parameters by minimizing the $L_1$ loss
function.

\begin{figure*}[!]
\centering
\subfigure[Noisy input]{\includegraphics[width=0.1063\textheight]{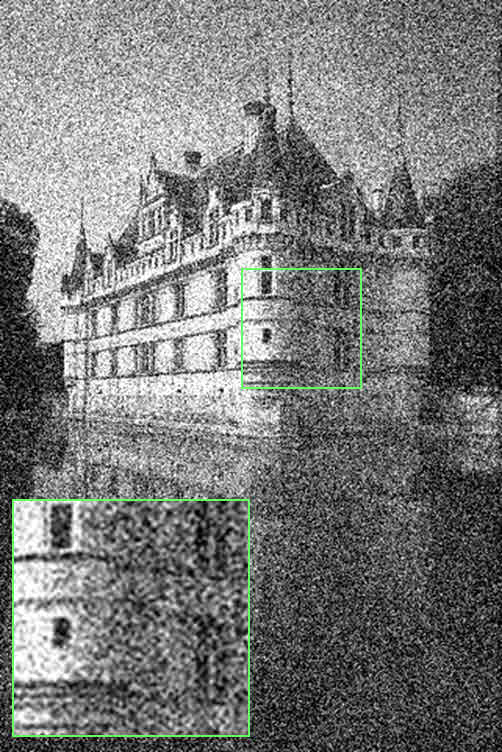}}
\subfigure[BM3D \cite{Dabov2007}]{\includegraphics[width=0.1063\textheight]{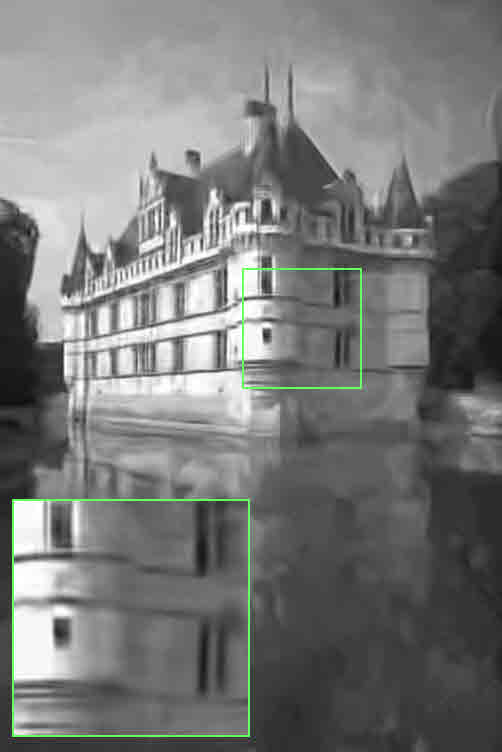}}
\subfigure[EPLL \cite{Zoran2011}]{\includegraphics[width=0.1063\textheight]{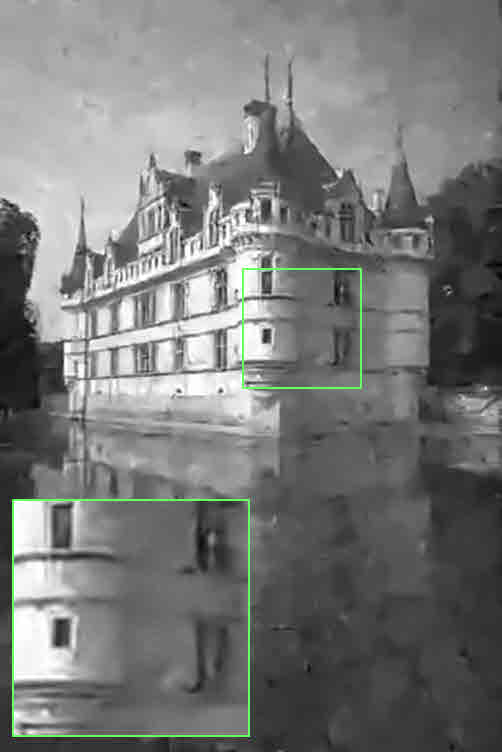}}
\subfigure[MLP \cite{Burger2012}]{\includegraphics[width=0.1063\textheight]{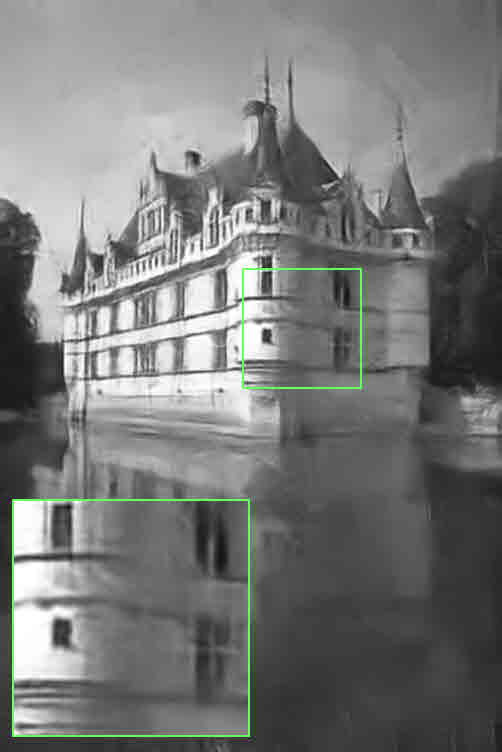}}
\subfigure[TRD \cite{Chen2015}]{\includegraphics[width=0.1063\textheight]{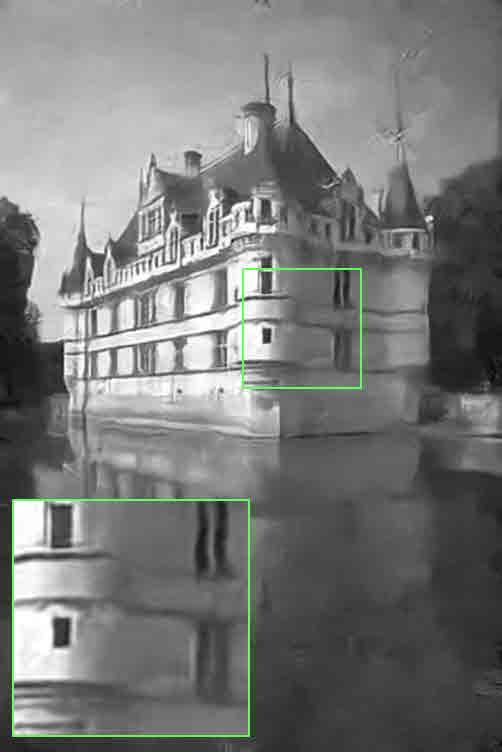}}
\subfigure[WNNM \cite{Gu2014}]{\includegraphics[width=0.1063\textheight]{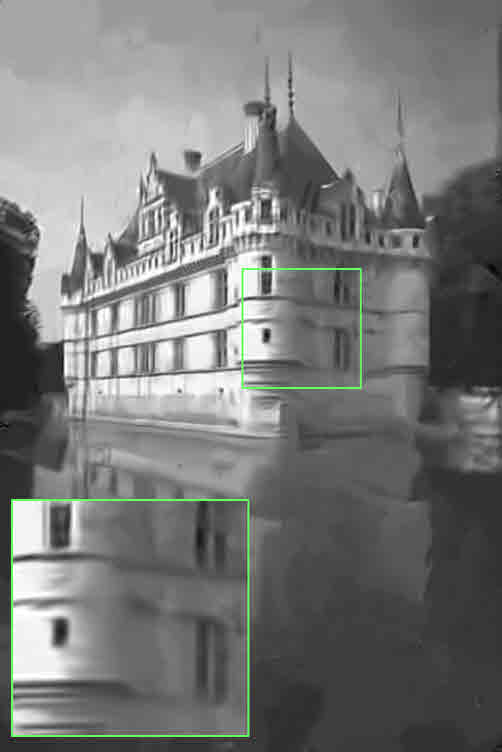}}
\subfigure[$\text{DeepAM}^{(3)}$]{\includegraphics[width=0.1063\textheight]{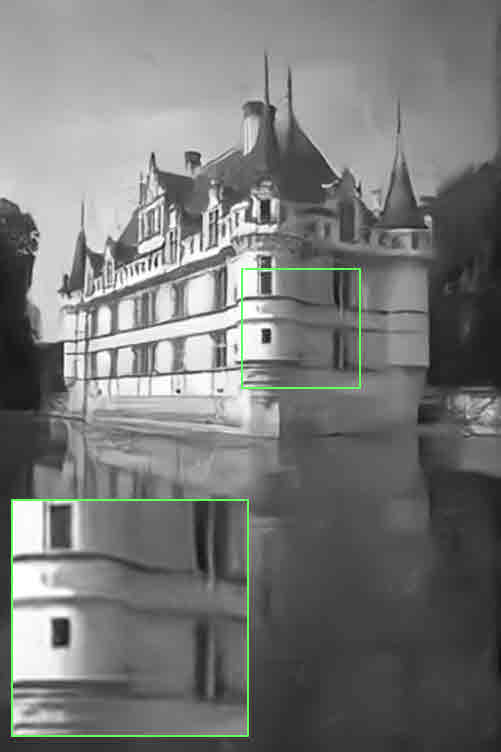}}\\
\vspace{-3pt}
\caption{Denoising examples with $\sigma=$50. (from left to right) noisy input, BM3D \cite{Dabov2007}, EPLL \cite{Zoran2011}, MLP \cite{Burger2012}, TRD \cite{Chen2015}, WNNM \cite{Gu2014}, and $\text{DeepAM}^{(3)}$. The input image is from the BSD68 \cite{Roth09}.}
\label{fig:ff}
\end{figure*}

\begin{table*}[t]
\centering
\begin{center}
\caption{The PSNR results on 12 images ($\sigma=25$). The CSF \cite{Schmidt2014} and TRD \cite{Chen2015} run 5 stages with $7\times7$ kernels.}
\label{table:aa}
\vspace{-3pt}
\resizebox{\textwidth}{!}{
\begin{tabular}{ccccccccccccccccccccc} \toprule
                                    & C. Man         & House           & Pepp.          & Starf.        & Fly           & Airpl.        & Parrot       & Lena         & Barb.       & Boat   & Man   & Couple\\\midrule\midrule
BM3D \cite{Dabov2007}               & 29.47          & 32.99           & 30.29          & 28.57         & 29.32         & 28.49         & 28.97        & 32.03        & 30.73        & 29.88  & 29.59 &29.70 \\
CSF  \cite{Schmidt2014}             & 29.51          & 32.41           & 30.32          & 28.87         & 29.69         & 28.80         & 28.91        & 31.87        & 28.99        & 29.75  & 29.68 &29.50 \\
EPLL \cite{Zoran2011}               & 29.21          & 32.14           & 30.12          & 28.48         & 29.35         & 28.66         & 28.96        & 31.58        & 28.53        & 29.64  & 29.57 &29.46 \\
MLP  \cite{Burger2012}              & 29.36          & 32.53           & 30.20          & 28.88         & 29.73         & 28.84         & 29.11        & 32.07        & 29.17        & 29.86  & 29.79 &29.68 \\
TRD  \cite{Chen2015}                & 29.71          & 32.62           & 30.57          & 29.05         & 29.97         & 28.95         & 29.22        & 32.02        & 29.39        & 29.91  & 29.83 &29.71 \\
WNNM \cite{Gu2014}                  & 29.63          & \textbf{33.39}  & 30.55          & 29.09         & 29.98         & 28.81         & 29.13        & 32.24        &\textbf{31.28}& 29.98  &29.74 & 29.80 \\
$\text{DeepAM}^{(3)}$                   & \textbf{29.97} & 33.35           & \textbf{30.89} & \textbf{29.43}& \textbf{30.27}&\textbf{29.03} &\textbf{29.41}&\textbf{32.52}& 29.52        & \textbf{30.23}&\textbf{30.07}&\textbf{30.15} \\
\bottomrule
\end{tabular}}
\end{center}
\vspace{-10pt}
\end{table*}

\begin{equation}
\label{eq:10}
\mathcal{L}= \frac{1}{M}\sum\limits_p {{{\lVert {u^{(p)} - {t^{(p)}}} \rVert}_1}},
\end{equation}
where $t^{(p)}$ and $u^{(p)}$ denote the ground truth image and the
output of the last reconstruction layer in (\ref{eq:7}), respectively. It is
known that $L_1$ loss in deep networks reduces splotchy artifacts
and outperforms $L_2$ loss for pixel-level prediction tasks
\cite{Zhao2015}. We use the stochastic gradient descent (SGD) to
minimize the loss function of (\ref{eq:10}). The derivative for the
back-propagation is obtained as follows:

\begin{equation}
\label{eq:11}
\frac{{\partial {\mathcal{L}^{(p)}}}}{{\partial {u^{(p)}}}} =
\text{sign}({u^{(p)}} - {t^{(p)}}).
\end{equation}
To learn the parameters in the network, we need the derivatives of
the loss ${\mathcal{L}^{(p)}}$ with respect to $v^{(p)}$ and
$\gamma^{(p)}$. By the chain rule of differentiation,
$\frac{{\partial {\mathcal{L}^{(p)}}}}{{\partial {v^{(p)}}}}$ can be
derived from (\ref{eq:7}):

\begin{equation}
\label{eq:12}
L\frac{{\partial {{\cal L}^{(p)}}}}{{\partial {v^{(p)}}}} = \left[ {{D_x}\frac{{\partial {{\cal L}^{(p)}}}}{{\partial {u^{(p)}}}},{D_y}\frac{{\partial {{\cal L}^{(p)}}}}{{\partial {u^{(p)}}}}} \right].
\end{equation}
$\frac{{\partial {\mathcal{L}^{(p)}}}}{{\partial {v^{(p)}}}}$ is obtained by solving the linear system of (\ref{eq:12}).
Similarly for $\frac{{\partial {\mathcal{L}^{(p)}}}}{{\partial {\gamma^{(p)}}}}$, we have:

\begin{equation}
\label{eq:13}
\frac{{\partial {\mathcal{L}^{(p)}}}}{{\partial {\gamma^{(p)}}}} = \left(L^{-1}\frac{{\partial {\mathcal{L}^{(p)}}}}{{\partial {u^{(p)}}}}\right) \circ ({f^{(p)}} - {u^{(p)}}),
\end{equation}
where ``$ \circ $" is an element-wise multiplication. Since the loss
${\mathcal{L}^{(p)}}$ is a scalar value, $\frac{{\partial
{\mathcal{L}^{(p)}}}}{{\partial {\gamma^{(p)}}}}$ and
$\frac{{\partial {\mathcal{L}^{(p)}}}}{{\partial {v^{(p)}}}}$ are
$N\times1$ and $N\times2$ vectors, respectively, where $N$ is total
number of pixels. More details about the derivations of (\ref{eq:12}) and (\ref{eq:13})
are available in the supplementary material. The system matrix $L$
is shared in (\ref{eq:12}) and (\ref{eq:13}), thus its incomplete factorization is performed only once.

Figure~\ref{fig:ee} shows the convergence of the PCG method for solving
the linear system of (\ref{eq:12}). We find that a few PCG iterations are
enough for the backpropagation. The average residual,
$\lVert L\frac{{\partial {\mathcal{L}^{(p)}}}}{{\partial v_x^{(p)}}} - {D_x}\frac{{\partial {\mathcal{L}^{(p)}}}}{{\partial {u^{(p)}}}} \rVert$
on 20 images is 1.3$\times$$10^{-6}$, after 10 iterations. The table
in Fig.~\ref{fig:ee} compares the runtime of PCG iterations and MATLAB
backslash (on 256$\times$256 image). The PCG with 10 iterations is about 5 times faster
than the direct linear system solver.

\section{Experiments}
We jointly train our DeepAM for 20 epochs.
From here on, we call $\text{DeepAM}^{(K)}$ the method trained through a cascade of $K$ $\text{DeepAM}$ iterations.
The MatConvNet library \cite{MatConvnet} (with 12GB NVIDIA Titan GPU) is used for network
construction and training. The networks are initialized randomly
using Gaussian distributions. The momentum and weight decay
parameters are set to 0.9 and 0.0005, respectively. We do not
perform any pre-training (or fine-tuning). The proposed method is
applied to single image denoising, depth super-resolution, and
RGB/NIR restoration. The results for the comparison with other
methods are obtained from source codes provided by the authors.
Additional results and analyses are available in the supplementary
material.

\begin{table*}[]
\centering
\caption{Average PSNR/SSIM on 68 images from \cite{Roth09} for image denoising with $\sigma=15,25,$ and $50$.}
\label{table:bb}
\begin{tabular}{cccccccc}
\toprule
\multirow{2}{*}{$\sigma$} & \multicolumn{7}{c}{PSNR / SSIM}     \vspace{4pt}                \\
                       & BM3D \cite{Dabov2007}   & MLP \cite{Burger2012} & CSF \cite{Schmidt2014} & TRD \cite{Chen2015} & $\text{DeepAM}^{(1)}$           & $\text{DeepAM}^{(2)}$               & $\text{DeepAM}^{(3)}$ \\
\midrule\midrule
15                     & 31.12 / 0.872          & -                  & 31.24 / 0.873         & 31.42 / 0.882            & 31.40 / 0.882               & 31.65 / 0.885                   & \textbf{31.68 / 0.886}     \\
25                     & 28.61 / 0.801          & 28.84  / 0.812     & 28.73 / 0.803         & 28.91 / 0.815            & 28.95 / 0.816               & 29.18 / 0.824                   & \textbf{29.21 / 0.825}     \\
50                     & 25.65 / 0.686          & 26.00  / 0.708     & -                     & 25.96 / 0.701            & 25.94 / 0.701               & 26.20 / 0.714                   & \textbf{26.24 / 0.716}     \\
\bottomrule
\end{tabular}
\vspace{-8pt}
\end{table*}

\begin{figure*}[!]
\centering
\renewcommand{\thesubfigure}{}
\subfigure{\includegraphics[width=0.125\textheight]{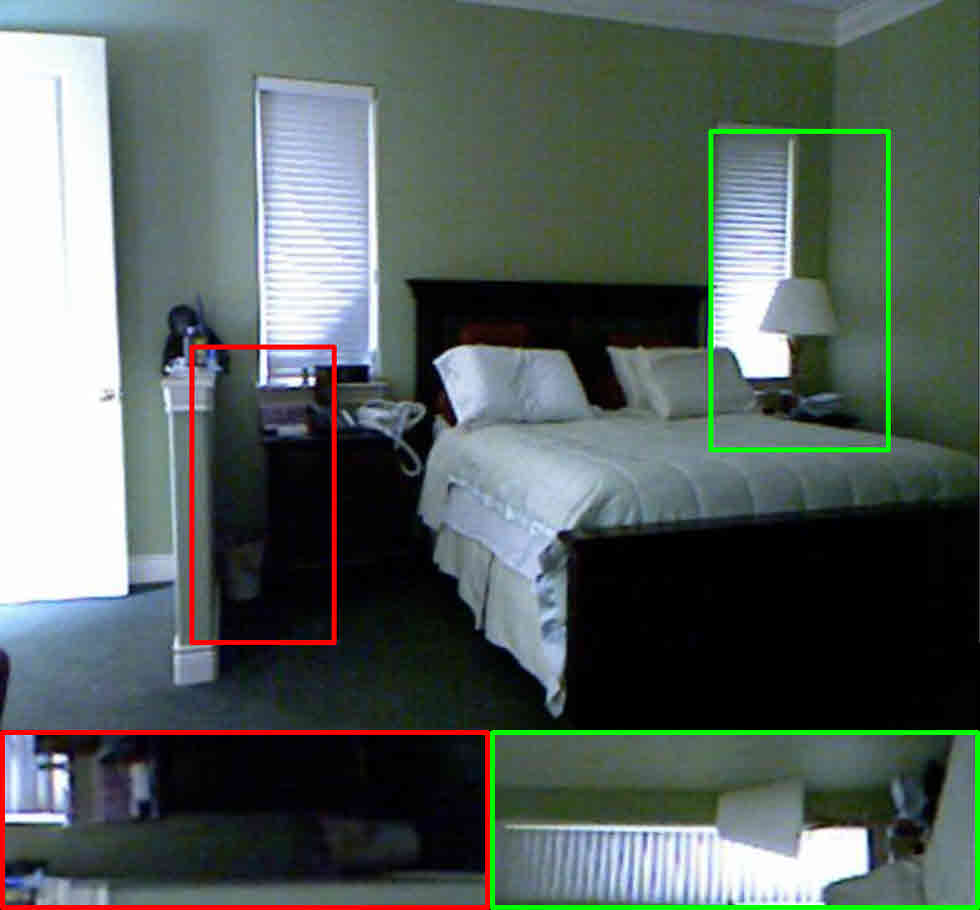}}
\subfigure{\includegraphics[width=0.125\textheight]{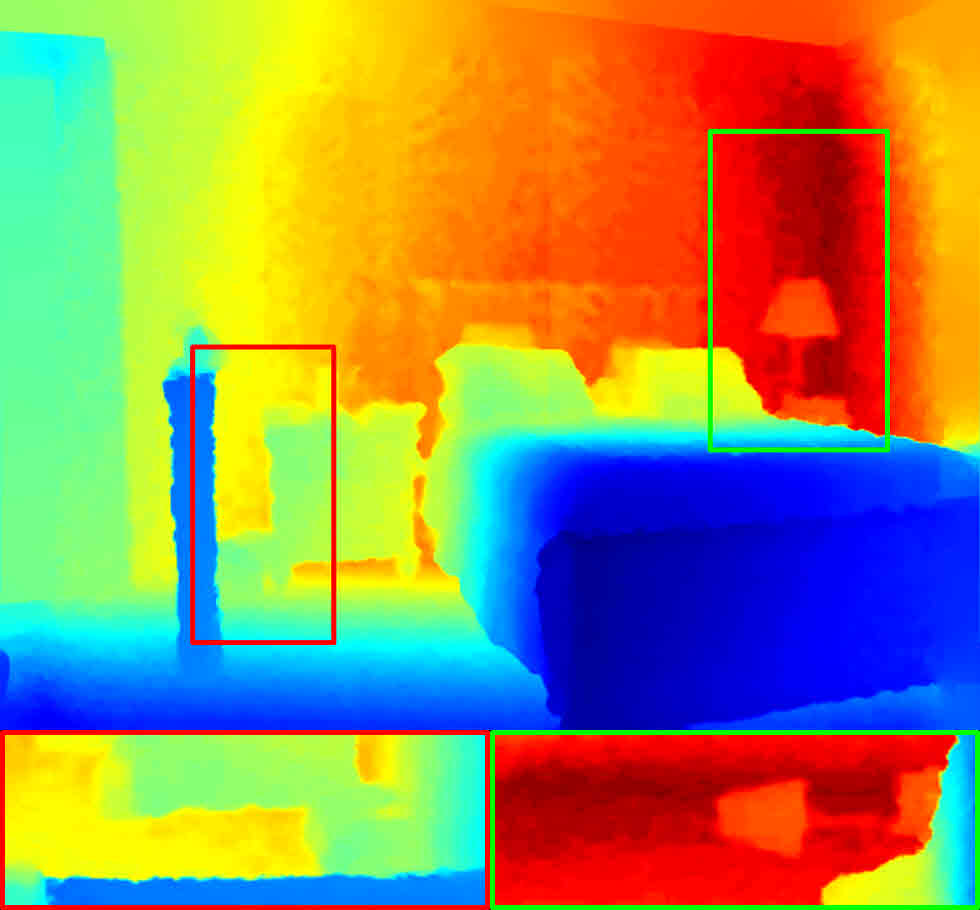}}
\subfigure{\includegraphics[width=0.125\textheight]{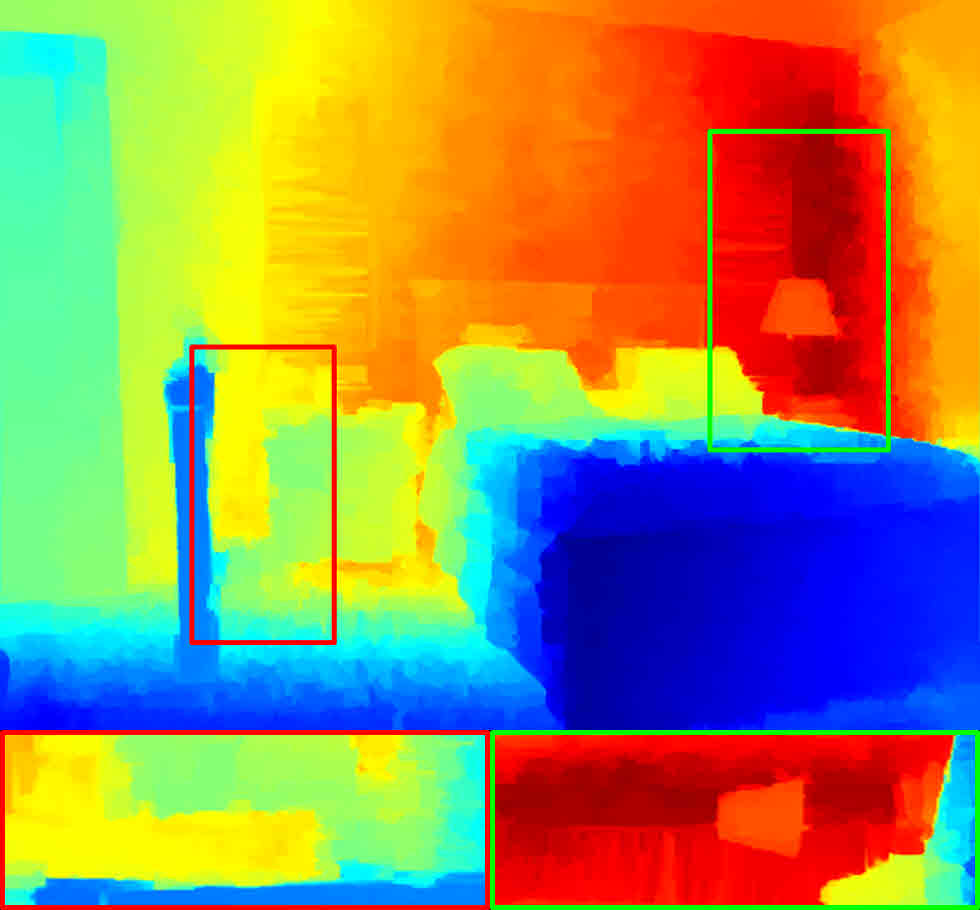}}
\subfigure{\includegraphics[width=0.125\textheight]{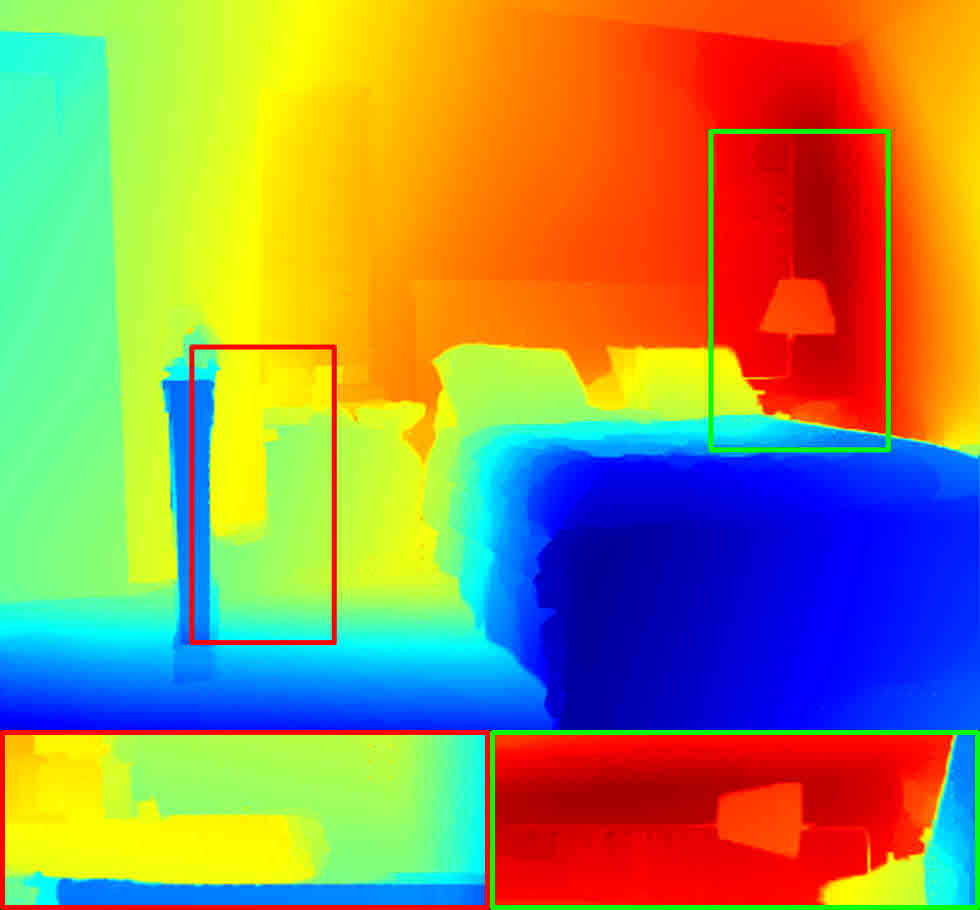}}
\subfigure{\includegraphics[width=0.125\textheight]{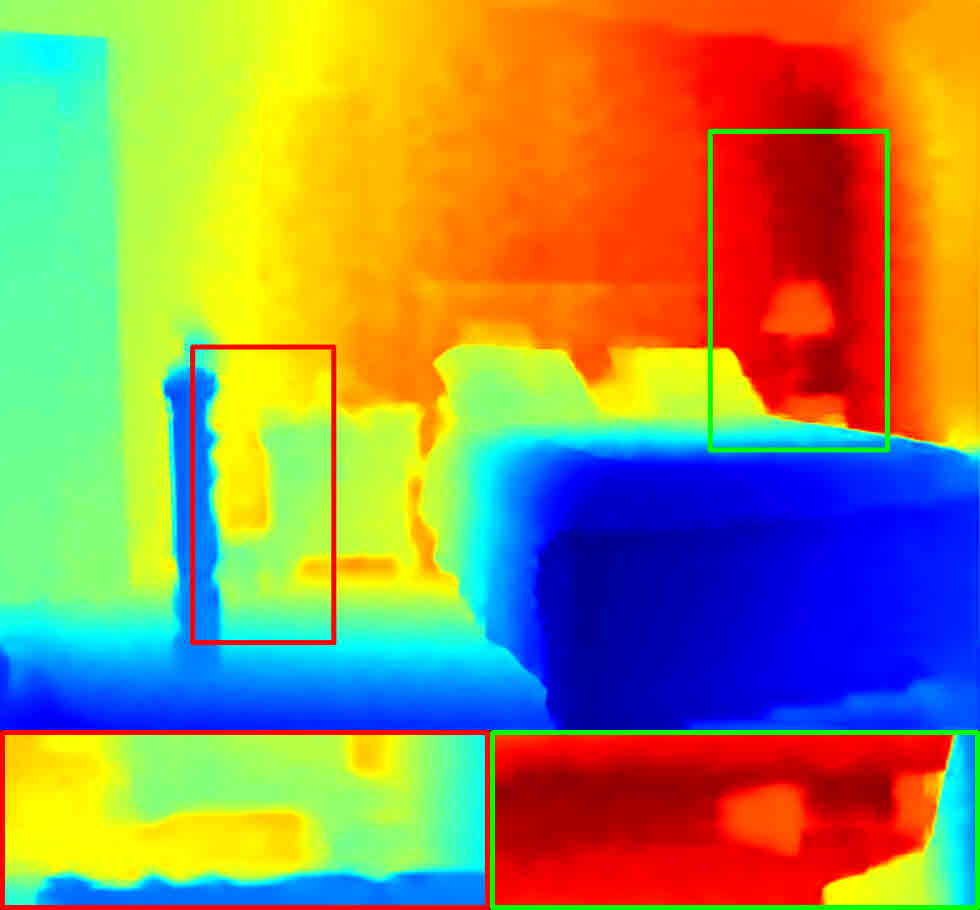}}
\subfigure{\includegraphics[width=0.125\textheight]{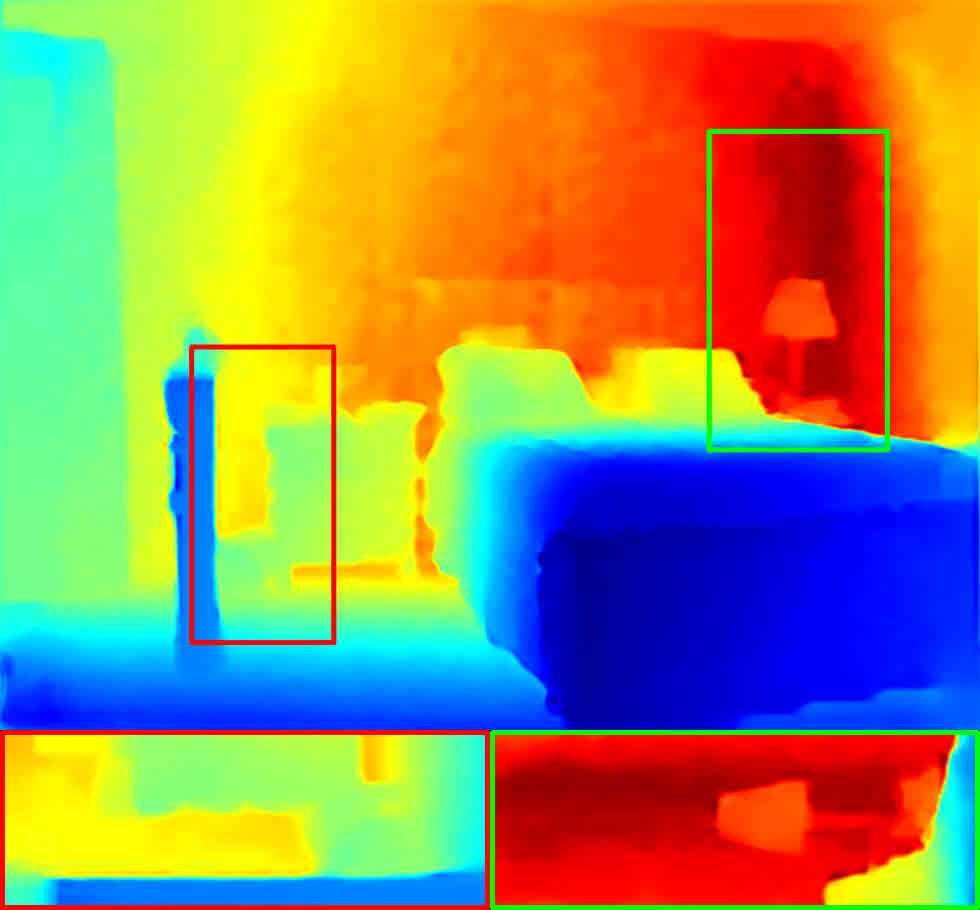}}\\
\vspace{-8pt}
\subfigure[(a) RGB image]            {\includegraphics[width=0.125\textheight]{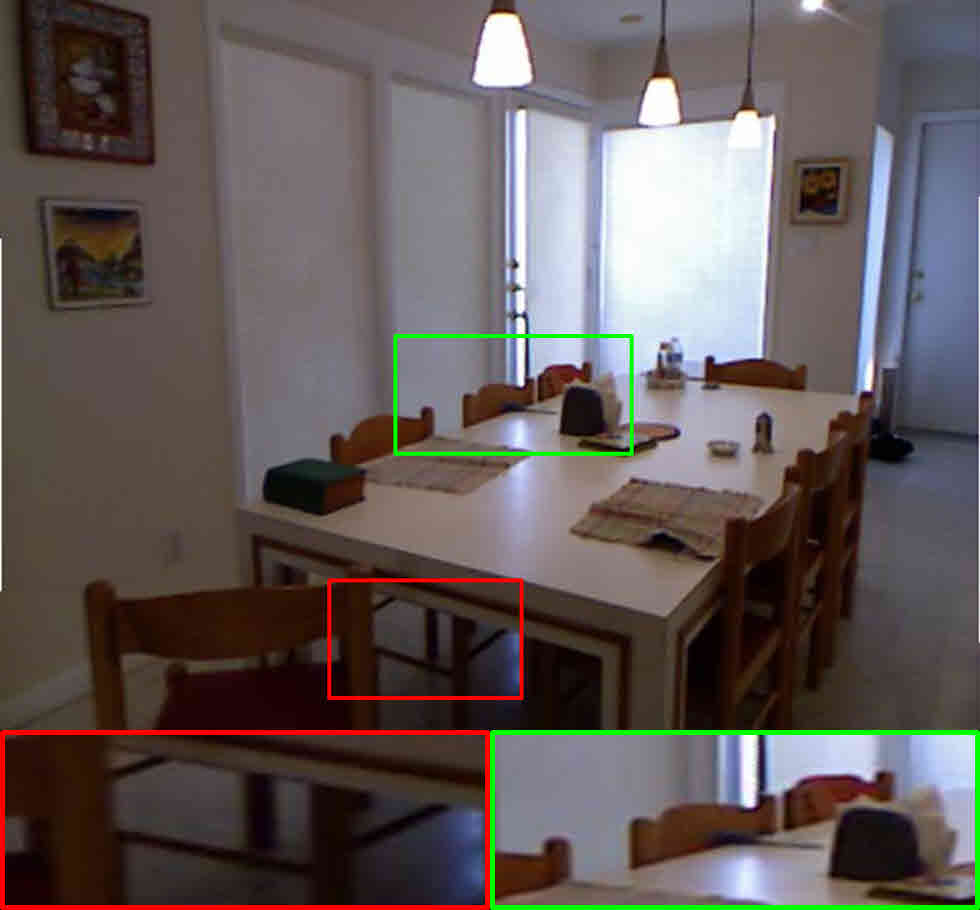}}
\subfigure[(b) ground truth]         {\includegraphics[width=0.125\textheight]{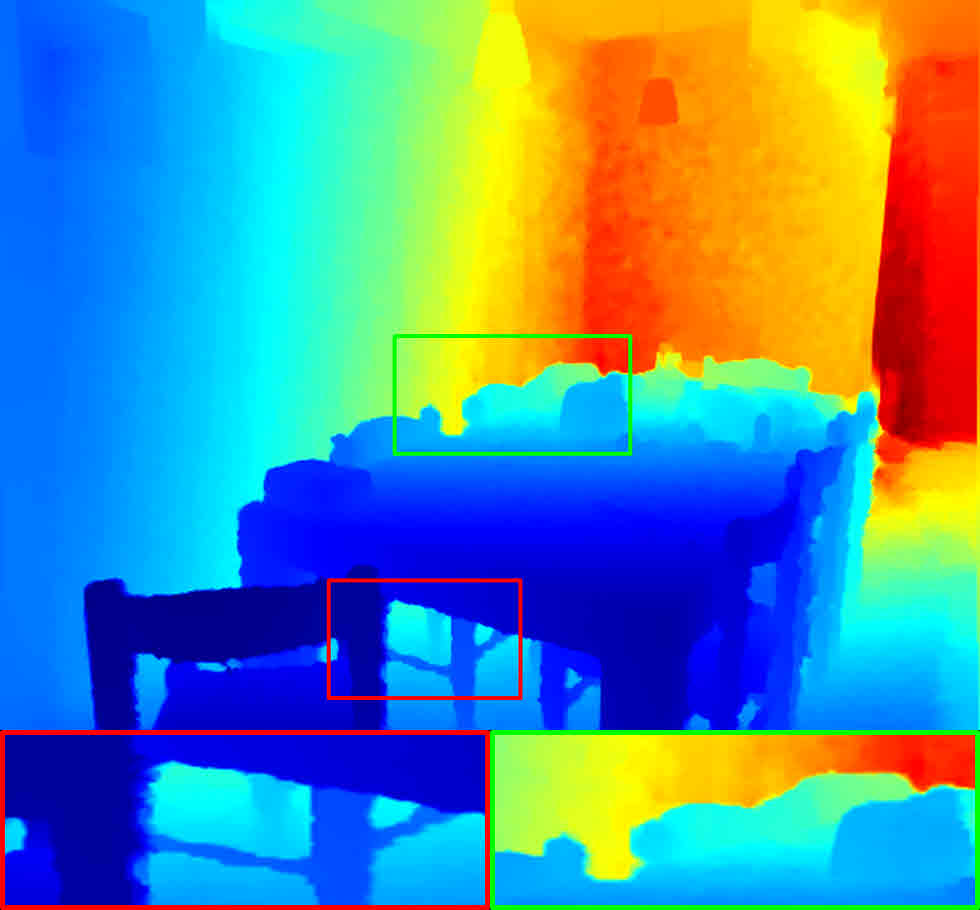}}
\subfigure[(c) NMRF \cite{Park2011}] {\includegraphics[width=0.125\textheight]{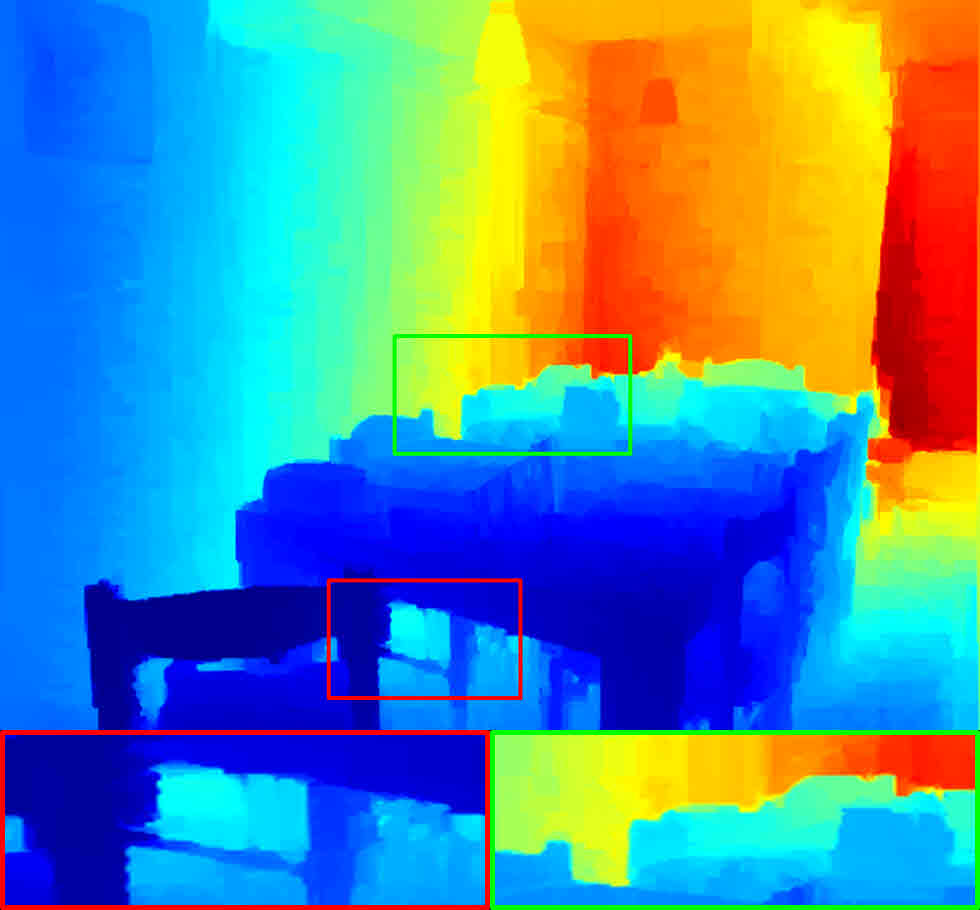}}
\subfigure[(d) TGV \cite{Ferstl2013}]{\includegraphics[width=0.125\textheight]{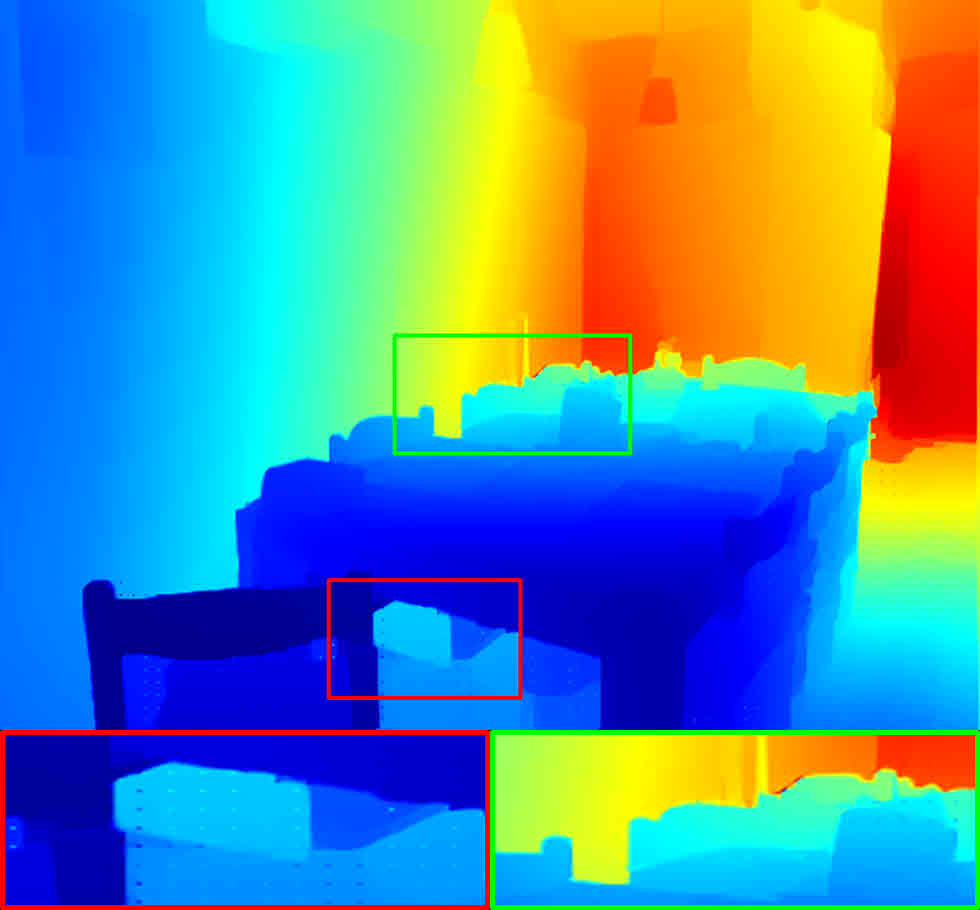}}
\subfigure[(e) DJF \cite{Li2016}]    {\includegraphics[width=0.125\textheight]{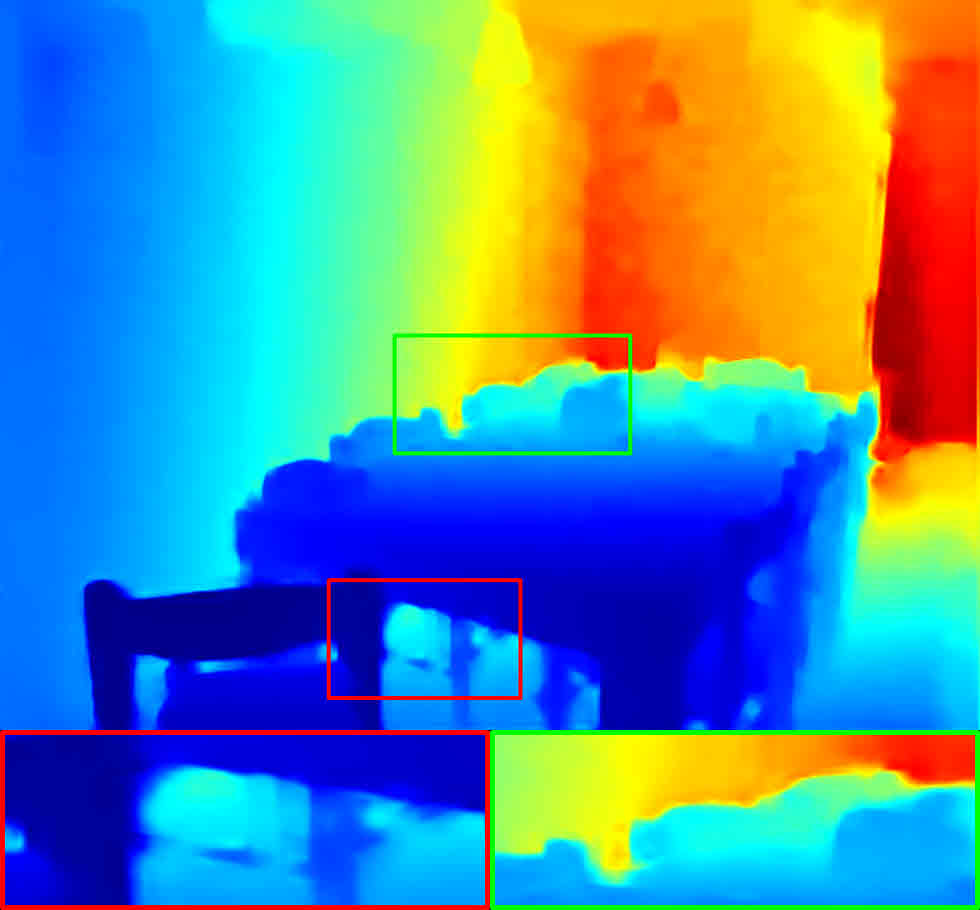}}
\subfigure[(f) $\text{DeepAM}^{(2)}$]    {\includegraphics[width=0.125\textheight]{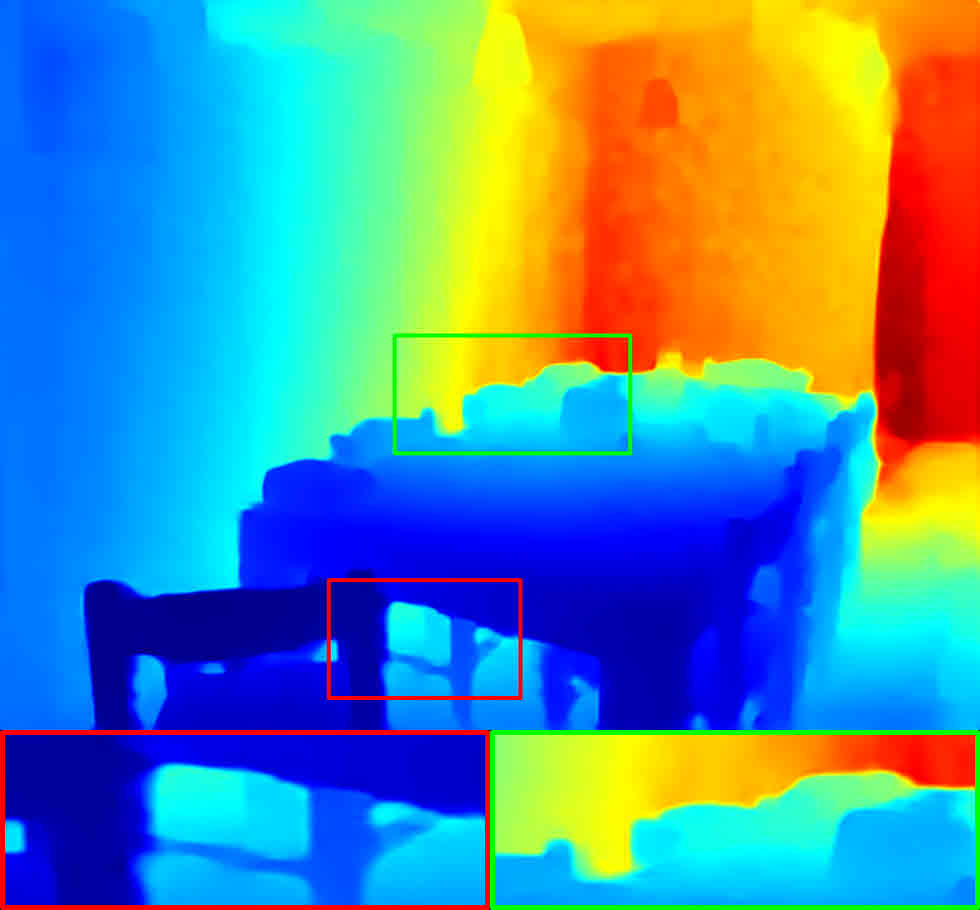}}\\
\vspace{-4pt}
\caption{Depth super-resolution examples ($\times8$): (a) RGB image, (b) ground truth, (c) NMRF \cite{Park2011}, (d) TGV \cite{Ferstl2013}, (e) DJF \cite{Li2016}, and (f) $\text{DeepAM}^{(2)}$. \vspace{-5pt}}
\label{fig:gg}
\end{figure*}

\subsection{Single Image Denoising}
We learned the $\text{DeepAM}^{(3)}$ from a set of $10^5$, $32\times32$
patches sampled from the BSD300 \cite{Martin2001} dataset.
Here $K$ was set to 3 as the performance of the DeepAM$^{(K)}$ converges
after 3 iterations (refer to Table~\ref{table:bb}). The noise levels were set to
$\sigma = 15, 25$, and $50$.
We compared against a variety of recent state-of-the-art techniques, including
BM3D \cite{Dabov2007}, WNNM \cite{Gu2014}, CSF \cite{Schmidt2014}, TRD \cite{Chen2015}, EPLL
\cite{Zoran2011}, and MLP \cite{Burger2012}.
The first two methods are based on the nonlocal regularization and the others are learning-based
approaches.

Table~\ref{table:aa} shows the peak signal-to-noise
ratio (PSNR) on the 12 test images \cite{Dabov2007}. The best results for each image
are highlighted in bold. The $\text{DeepAM}^{(3)}$ yields the highest
PSNR results on most images.
We could find that our deep aggregation used in the mapping step outperforms the point-wise mapping of the CSF \cite{Schmidt2014} by 0.3$\thicksim$0.5dB.
Learning-based methods tend to have better performance than
handcrafted models. We, however, observed that the methods (BM3D
\cite{Dabov2007} and WNNM \cite{Gu2014}) based on the nonlocal
regularization usually work better on images that are dominated by
repetitive textures, e.g., `House' and `Barbara'.
The nonlocal self-similarity is a powerful prior on regular and
repetitive texture, but it may lead to inferior results on irregular
regions.

\begin{table}[]
\centering
\caption{Average BMP ($\delta=$3) on 449 images from the NYU v2 dataset \cite{Nathan2012} and on 10 images from the Middlebury dataset \cite{Scharstein2002}. Depth values are normalized within
the range [0,255].}
\label{table:cc}
\resizebox{\textwidth}{!}{
\begin{tabular}{cccc}
\toprule
                           & \multicolumn{3}{c}{BMP ($\delta=$3):  NYU v2 \cite{Nathan2012} / Middlebury \cite{Scharstein2002}} \vspace{4pt}\\
Method                     & $\times4$        & $\times8$        & $\times16$       \\
\midrule\midrule
NMRF      \cite{Park2011}  & 1.41          / 4.56          &  4.21          / 7.59              &  16.25         / 13.22             \\
TGV       \cite{Ferstl2013}& 1.58          / 5.72          &  5.42          / 8.82              &  17.89         / 13.47             \\
SD filter \cite{Ham2015}   & 1.27          / \textbf{2.41} &  3.56          / 5.97              &  15.43         / 12.18             \\
DJF       \cite{Li2016}    & 0.68          / 3.75          &  1.92          / 6.37              &  5.82          / 12.63             \\
$\text{DeepAM}^{(2)}$          & \textbf{0.57} / 3.14          &  \textbf{1.58} / \textbf{5.78}     &  \textbf{4.63} / \textbf{10.45}    \\
\bottomrule
\end{tabular}}
\vspace{-10pt}
\end{table}

Figure~\ref{fig:ff} shows denoising results using one image from the BSD68 dataset \cite{Roth09}.
The $\text{DeepAM}^{(3)}$ visually outperforms state-of-the-art methods.
Table~\ref{table:bb} summarizes an objective evaluation by measuring average PSNR and structural similarity indexes (SSIM) \cite{Wang04}
on 68 images from the BSD68 dataset \cite{Roth09}.
As expected, our method achieves a significant improvement over the nonlocal-based method as well as the recent data-driven approaches.
Due to the space limit, some methods were omitted in the table, and full performance comparison is available in the supplementary materials.

\subsection{Depth Super-resolution}
Modern depth sensors, e.g. MS Kinect, provide dense depth
measurement in dynamic scene, but typically have a low resolution. A
common approach to tackle this problem is to exploit a
high-resolution (HR) RGB image as guidance. We applied our
$\text{DeepAM}^{(2)}$ to this task, and evaluated it on the NYU v2 dataset \cite{Nathan2012} and Middlebury dataset \cite{Scharstein2002}.
The NYU v2 dataset \cite{Nathan2012} consists of 1449 RGB-D image pairs of
indoor scenes, among which 1000 image pairs were used for training
and 449 image pairs for testing. Depth values are normalized within
the range [0,255]. To train the network, we randomly collected $10^5$
RGB-D patch pairs of size $32\times32$ from training set. A
low-resolution (LR) depth image was synthesized by nearest-neighbor
downsampling ($\times4$, $\times8$, and $\times16$). The network
takes the LR depth image, which is bilinearly interpolated into the
desired HR grid, and the HR RGB image as inputs.

\begin{figure}[!]
\centering
\subfigure[IR image]{\includegraphics[width=0.1205\textheight]{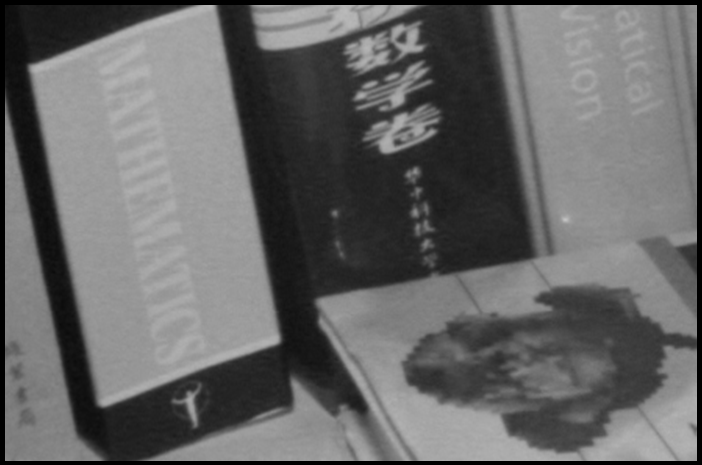}}
\subfigure[RGB image]{\includegraphics[width=0.1205\textheight]{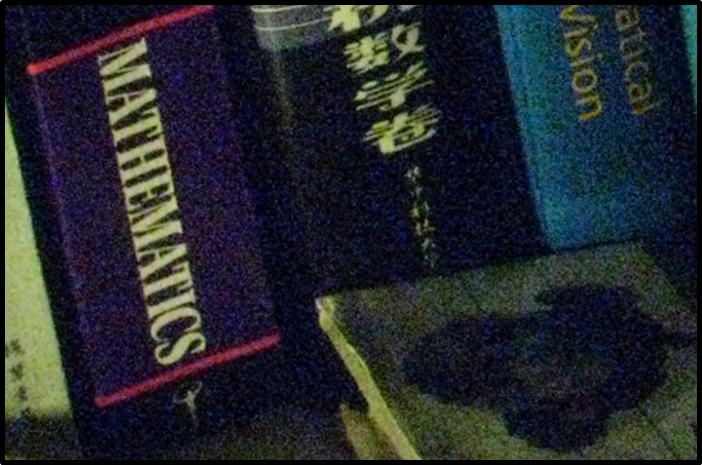}}
\subfigure[Cross-field \cite{Shen2015}]{\includegraphics[width=0.1205\textheight]{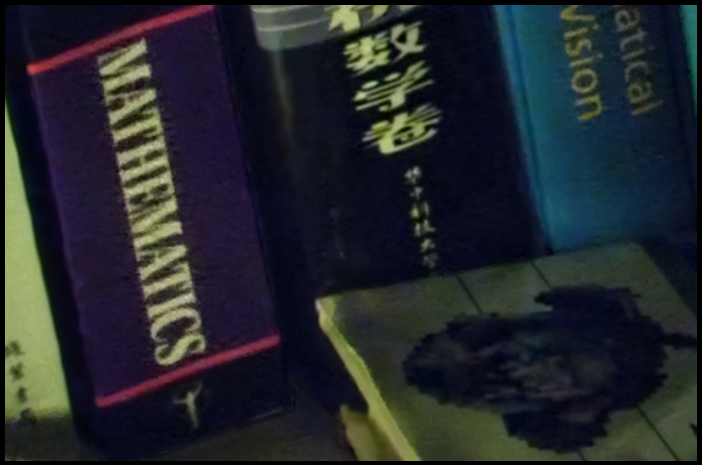}}\\
\vspace{-3pt}
\subfigure[DJF \cite{Li2016}]{\includegraphics[width=0.1205\textheight]{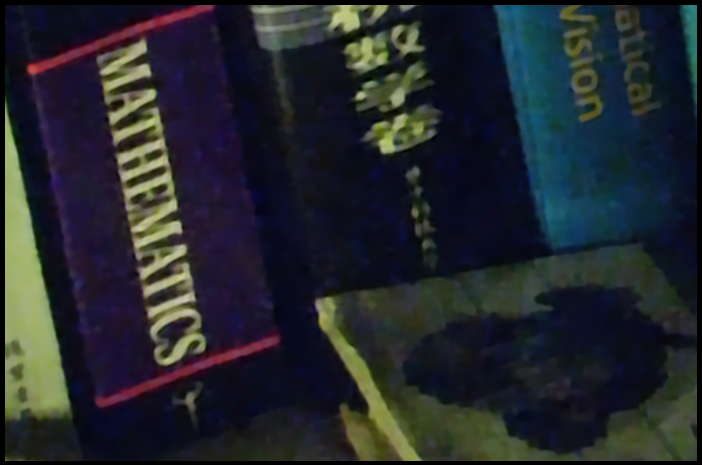}}
\subfigure[$\text{DeepAM}^{(2)}_{\sigma=25}$]{\includegraphics[width=0.1205\textheight]{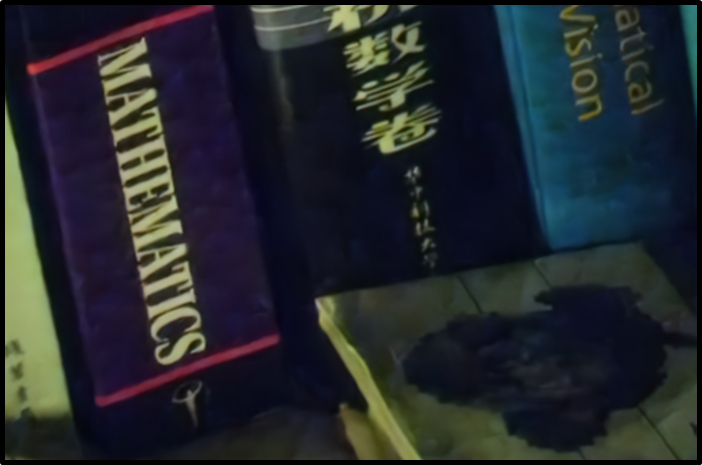}}
\subfigure[$\text{DeepAM}^{(2)}_{\sigma=50}$]{\includegraphics[width=0.1205\textheight]{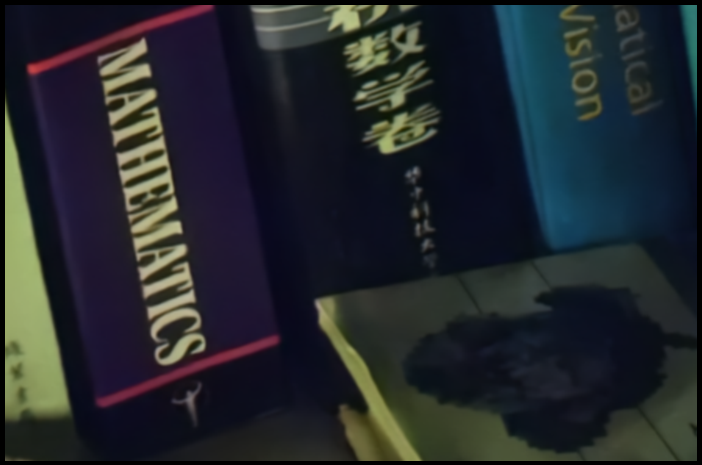}}\\
\vspace{-5pt}
\caption{RGB/NIR restoration for real-world examples: (a) RGB image, (b) NIR image, (c) Cross-field \cite{Shen2015}, (d) DJF \cite{Li2016}, (e) $\text{DeepAM}^{(2)}$ trained with $\sigma=25$, and (f) $\text{DeepAM}^{(2)}$ trained with $\sigma=50$. The result of (c) is from the project webpage of \cite{Shen2015}.
\label{fig:hh}
\vspace{-10pt}}
\end{figure}

Figure~\ref{fig:gg} shows the super-resolution results of NMRF
\cite{Park2011}, TGV \cite{Ferstl2013}, deep joint image filtering
(DJF) \cite{Li2016}, and $\text{DeepAM}^{(2)}$. The TGV model
\cite{Ferstl2013} uses an anisotropic diffusion tensor that solely
depends on the RGB image. The major drawback of this approach is
that the RGB-depth coherence assumption is violated in textured surfaces. Thus, the restored depth image may
contain gradients similar to the color image, which causes texture
copying artifacts (Fig.~\ref{fig:gg}(d)). Although the NMRF
\cite{Park2011} combines several weighting schemes, computed from RGB
image, segmentation, and initially interpolated depth, the texture
copying artifacts are still observed (Fig.~\ref{fig:gg}(c)). The
NMRF \cite{Park2011} preserves depth discontinuities well, but shows
poor results in smooth surfaces. The DJF \cite{Li2016} avoids the texture copying
artifacts thanks to faithful CNN responses extracted from both color
image and depth map (Fig.~\ref{fig:gg}(e)). However, this
method lacks the regularization constraint that
encourages spatial and appearance consistency on the output, and
thus it over-smooths the results and does not protect thin
structures. Our $\text{DeepAM}^{(2)}$ preserves sharp depth
discontinuities without notable artifacts as shown in
Fig.~\ref{fig:gg}(f). The quantitative evaluations on the NYU v2 dataset \cite{Nathan2012} and Middlebury dataset \cite{Scharstein2002} are summarized in
Table~\ref{table:cc}. The accuracy is measured by the bad matching percentage (BMP) \cite{Scharstein2002} with
tolerance $\delta=$3.

\subsection{RGB/NIR Restoration}
The RGB/NIR restoration aims to enhance a noisy RGB image taken
under low illumination using a spatially aligned NIR image. The
challenge when applying our model to the RGB/NIR
restoration is the lack of the ground truth data for training.
For constructing a large training data, we used the indoor IVRL
dataset consisting of 400 RGB/NIR pairs \cite{Salamati2014} that were recorded under daylight illumination\footnote{This dataset \cite{Salamati2014} was originally introduced for semantic segmentation.}.
Specifically, we generated noisy RGB images by adding the synthetic Gaussian noise
with $\sigma=25$ and $50$, and used 300 image pairs for training.

In Table~\ref{table:dd}, we performed an objective evaluation using 5 test images in \cite{Honda2015}.
The $\text{DeepAM}^{(2)}$ gives better quantitative results than other state-of-the-art methods \cite{Shen2015,Dabov2007,Ham2015}.
Figure~\ref{fig:hh} compares the RGB/NIR restoration results of Cross-field \cite{Shen2015}, DJF \cite{Li2016}, and our $\text{DeepAM}^{(2)}$ on the real-world example.
The input RGB/NIR pair was taken from the project website of \cite{Shen2015}.
This experiment shows the proposed method can be applied to real-world data, although it was trained from the synthetic dataset.
It was reported in \cite{Honda2015} that the
restoration algorithm designed (or trained) to work under a daylight
condition could also be used for both daylight and night conditions.

\begin{table}[]
\centering
\caption{The PSNR results with 5 RGB/NIR pairs from \cite{Honda2015}. The noisy RGB images are generated by adding the synthetic Gaussian noise.
\vspace{-5pt}}
\label{table:dd}
\renewcommand{\thesubfigure}{}
\subfigure[$\sharp$1]{\includegraphics[width=0.07\textheight]{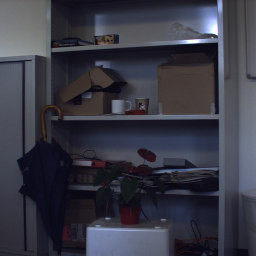}}
\subfigure[$\sharp$2]{\includegraphics[width=0.07\textheight]{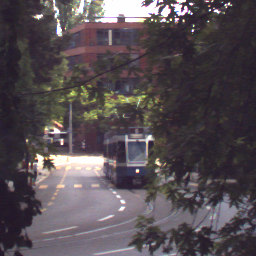}}
\subfigure[$\sharp$3]{\includegraphics[width=0.07\textheight]{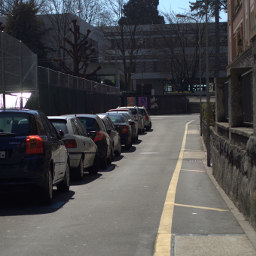}}
\subfigure[$\sharp$4]{\includegraphics[width=0.07\textheight]{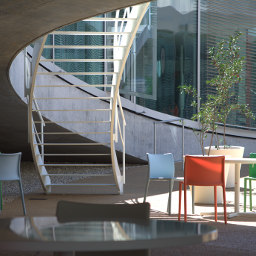}}
\subfigure[$\sharp$5]{\includegraphics[width=0.07\textheight]{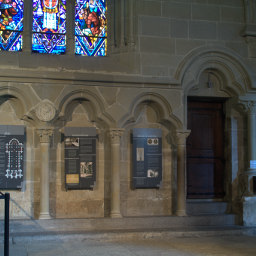}}\\
\vspace{-8pt}
\subfigure{\includegraphics[width=0.07\textheight]{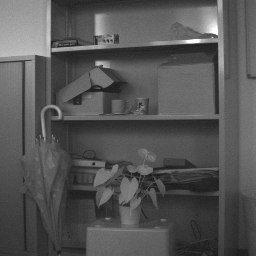}}
\subfigure{\includegraphics[width=0.07\textheight]{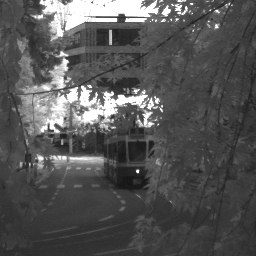}}
\subfigure{\includegraphics[width=0.07\textheight]{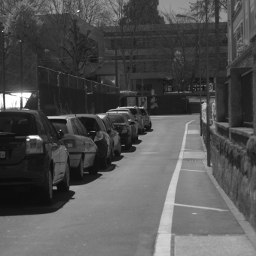}}
\subfigure{\includegraphics[width=0.07\textheight]{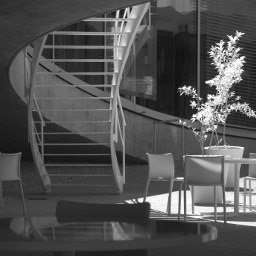}}
\subfigure{\includegraphics[width=0.07\textheight]{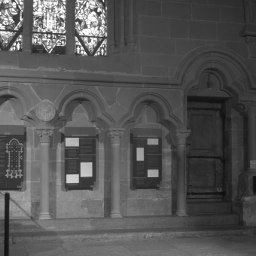}}
\vspace{3pt}
\resizebox{\textwidth}{!}{
\begin{tabular}{ccccc}
\toprule
                           & \multicolumn{4}{c}{PSNR} \vspace{4pt} \\
$\sigma=50$                       & BM3D \cite{Dabov2007}   & SD filter \cite{Ham2015}  & Cross-field \cite{Shen2015}            & $\text{DeepAM}^{(2)}$   \\
\midrule\midrule
Sequence 1                        & 31.86                   &  30.97        &  31.45                                 &      \textbf{32.84}   \vspace{3pt} \\
Sequence 2                        & 27.62                   &  26.13        &  27.59                                 &      \textbf{28.10}   \vspace{3pt} \\
Sequence 3                        & 28.08                   &  28.06        &  28.47                                 &      \textbf{30.43}   \vspace{3pt} \\
Sequence 4                        & 26.85                   &  25.65        &  26.91                                 &      \textbf{28.13}   \vspace{3pt} \\
Sequence 5                        & 26.52                   &  26.11       &  \textbf{26.98}                        &      26.94             \vspace{3pt} \\
\midrule\midrule
Average                           & 28.19                   &  27.38       &  28.28                       &      \textbf{29.28}              \\
\bottomrule
\end{tabular}}
\vspace{-10pt}
\end{table}

\section{Conclusion}
We have explored a general framework called the DeepAM,
which can be used in various image restoration applications.
Contrary to existing data-driven approaches that just produce the
restoration result from the CNNs,
the DeepAM uses the CNNs to learn the regularizer of the AM algorithm.
Our formulation fully integrates the CNNs with an energy minimization model, making it possible to learn whole networks in an end-to-end manner.
Experiments demonstrate that the deep aggregation in the mapping step is the critical factor of the proposed learning model.
As future work, we will further investigate an adversarial loss in pixel-level prediction tasks.

{\small
\bibliographystyle{ieee}
\bibliography{egbib}
}

\end{document}